\documentclass[11pt]{article}

\usepackage[margin=1in]{geometry}
\usepackage{graphicx}
\usepackage{booktabs}
\usepackage{microtype}
\usepackage{amsmath}
\usepackage{amssymb}
\usepackage[round,sort&compress]{natbib}
\usepackage[hidelinks]{hyperref}
\usepackage{caption}
\usepackage{xcolor}
\usepackage{setspace}
\usepackage{float}
\usepackage{array}
\usepackage{longtable}
\usepackage{placeins}

\graphicspath{{arxiv_assets/figures/}}
\definecolor{linkblue}{HTML}{0066A1}
\hypersetup{colorlinks=true,citecolor=linkblue,linkcolor=linkblue,urlcolor=linkblue}
\newcolumntype{P}[1]{>{\raggedright\arraybackslash}p{#1}}

\newif\ifblind
\ifdefined\ISRBlind\blindtrue\else\blindfalse\fi

\title{ Designing Behavioral Profiles for Language Model Simulation of Out-of-Distribution Human Behavior}
\ifblind
  \author{}
\else
  \author{Robin Na$^{1}$ \quad Kehang Zhu$^{2}$ \quad Logan Cross$^{3}$ \quad Michiel A. Bakker$^{1}$\\[0.6em]
  \small $^{1}$Massachusetts Institute of Technology\\
  \small $^{2}$Digital Economy Lab,  Stanford University\\
  \small $^{3}$Google DeepMind}
\fi
\date{}

\bibpunct[, ]{(}{)}{;}{a}{}{,}

\newcolumntype{C}[1]{>{\centering\arraybackslash}p{#1}}

\begin{document}
{\setstretch{1.0}\maketitle}

\begin{abstract}
\normalsize
Large language models (LLMs) are increasingly used to simulate human behavior, but which descriptions of people improve prediction when experimental designs change remains unclear. We develop a framework that draws candidate behavioral features from the literature and constructs profiles from earlier human experiments. An LLM-based profile generator assigns feature values from observed actions, messages, and surveys; a separate LLM simulates behavior using these profiles. By comparing simulated and observed outcomes in the training data, we select profile contents through forward feature selection, adding features while they improve prediction. We then evaluate the selected contents without modification against outcomes from new participants under new designs. Across 616 public-goods games, 144 multi-party bargaining games, and 1,480 generalized 11--20 money-request designs, leading one-feature profiles predict human outcome distributions better than no profile or general-purpose personas. Profiles with more features provide no clear additional benefit, and features' relative predictive value largely persists across learning and validation. Using profiles from one game family to simulate another produces uneven gains and can make predictions less accurate than simulations without profiles. The framework makes the choice of behavioral information a testable part of simulation design, helping researchers assess which descriptions remain useful as experimental conditions change.
\end{abstract}

\section{Introduction}

Large language models (LLMs) are widely used to simulate how people behave in social and economic settings: to anticipate how a population would respond to a rule that has not yet been tested and to screen interventions before running them on people \citep{Ashokkumar2026,ManningZhuHorton2024}. Recent studies also examine how AI systems change outcomes when they participate in or mediate human group tasks \citep{ShiradoChristakis2017,TesslerEtAl2024,DellAcquaEtAl2026}. Because the settings of greatest interest are often ones that have not been tested, the usefulness of such a simulation depends on out-of-distribution generalization: whether it continues to match the target population after the rules, incentives, or available actions change \citep{HullmanEtAl2026}.

A common way to improve LLM simulation is to condition the model on a description of the people it represents \citep{Argyle2023,CrossHaberYamins2025,ManningHorton2025}. Such descriptions can be built from survey backstories \citep{Argyle2023,ToubiaEtAl2025}, from interviews \citep{ParkEtAl2026}, or from records of earlier behavior \citep{ManningHorton2025}, and synthetic libraries of human profiles provide general-purpose \emph{personas} intended to provide diverse descriptions for simulation and synthetic-data generation \citep{GeEtAl2024,NVIDIANemotron2025,PaglieriEtAl2026}.

It is, however, much less clear how to construct such behavioral descriptions in ways that best simulate human behavior in new settings. A person's behavioral record can imply many different characteristics about them, and which of them is informative depends on the decision being simulated: a person's political ideology may be predictive of their voting behavior but much less so of their favorite color. Moreover, a richer description is not necessarily better. Models given demographic descriptions can exaggerate differences between groups, compress variation within them, or fall back on stereotypical associations when attributes conflict \citep{Bisbee2024,LiuEtAl2024,ChengEtAl2023}. Also, increasingly detailed personas can make simulated agents less distinct from one another \citep{BhattacharyyaEtAl2026}. The description in the simulator's prompt therefore becomes an information design problem: which \emph{behavioral features} should it include, and which should it leave out?

Here we develop a framework for designing \emph{behavioral profiles} from human experimental data to improve LLM simulations under new experimental designs (Figure~\ref{fig:design}). The framework has two stages, each using data from a different set of participants and experimental designs. In the \emph{profile learning stage}, we use human experimental data (the \emph{learning data}) to construct profiles and select their contents according to their downstream simulation performance. In the \emph{validation stage}, we evaluate the chosen profile contents on data from new participants facing designs absent from the learning data. In policy evaluation, for example, this could mean learning from experiments under an existing incentive program and validating predictions under revised payment rules; in market design, it could mean learning from bidding under one auction format and validating predictions under another.

We draw candidate behavioral features from the relevant theoretical and experimental literature and specify each feature's definition and scoring rules. Using the learning data, we then prompt an LLM to evaluate behavioral records containing observed actions and, where available, messages and survey responses. It assigns a value from 0 to 10 only when the record provides direct evidence for a feature and reports that evidence alongside the value. The resulting \emph{profile library} contains one evidence-grounded profile for each participant in the experiment. We call a profile containing information about every candidate feature a \emph{complete profile}.

We then sample profiles from this library and include them in the prompts used by a separate LLM to simulate human-subject experiments. For each learning comparison, the sampled profiles come from records separate from the human outcomes used to evaluate those simulations. Holding the simulated games and sampled profiles constant, we compare profiles containing different subsets of behavioral features, including individual features and larger combinations. We compare simulated and human behavior to identify which profile contents improve prediction. The individual feature that most consistently improves this match becomes the \emph{leading feature} and defines the \emph{leading one-feature profile}. We then use forward feature selection to add features one at a time while the expanded profile continues to improve the simulation \citep{GuyonElisseeff2003}. In the validation stage, we simulate the experimental designs represented in the validation data using the profile contents compared during profile learning. Comparing the simulated and observed outcomes then tests whether the leading profiles generalize to new participants and experimental designs.

We apply the framework separately to three \emph{game families}, each organized around a common type of strategic interaction and outcome. Public-goods games study repeated decisions about contributing to a shared return under different opportunities for communication, reward, and punishment \citep{AlsobayEtAl2026}. Multi-party bargaining studies negotiated exchanges among participants with privately known asset values \citep{QianEtAl2025}. Generalized 11--20 money-request games study one-shot strategic choices under varying payoff rules \citep{ManningHorton2025}. Together they provide 3,691 public-goods participant records, 144 bargaining participants, and 1,670 money-request choices for building profiles, and their validation data contain 250 public-goods games, 48 bargaining games, and 2,502 money-request choices (Table~\ref{tab:study-overview}). The three differ in social structure, information, available actions, and incentives, each has a mature literature from which to define behavioral features, and in each the learning and validation data contain different participants and nonoverlapping designs.

In all three families, profiles constructed this way generated LLM simulations that matched the human outcome distributions in validation more closely than simulations not conditioned on any description or those conditioned on general-purpose persona libraries from earlier studies (Figure~\ref{fig:heldout}). Complete profiles containing every identifiable candidate feature provided no clear advantage over the leading one-feature profiles. Individual features differed substantially in their predictive value: some reduced the prediction error considerably, others produced no improvement, and several left the simulation further from human behavior than simulation without a behavioral profile. These relative rankings stayed consistent between profile learning and validation (Figure~\ref{fig:transfer}). A separate analysis asked whether profiles constructed from one family could help simulate another (Figure~\ref{fig:cross-family}). Profiles generalized more successfully to new designs within the same game family than to a different type of game. In every target family, the leading one-feature profile learned from that family's own records achieved the highest distributional alignment. Cross-family transfer also depended on direction: an improvement over no profile in one direction could accompany a decline in the reverse comparison. Because the studies involve different participants between learning and validation data, our evaluation concerns population-level and experimental design-level transfer rather than same-person prediction across games.

By treating behavioral-profile construction as an information design problem, this study makes three contributions. First, it provides a framework for constructing interpretable behavioral profiles from unstructured records and selecting their contents using earlier experiments and evaluating them under new experimental designs. We build on \citet{PaglieriEtAl2026}, who organize persona generation around explicit axes of variation, by making profile-construction choices empirical. LLM-assisted literature synthesis and evidence extraction from human records systematize choices that would otherwise require manual configuration. We then determine which features, and how many, improve simulation, and evaluate the resulting profiles with new participants under new experimental designs. This contributes to design-oriented research on AI by showing that, in a predictive simulation system, the information used to represent people is itself a component to be designed and evaluated under the explicit goal of improving predictions \citep{Abbasi2024}.
Second, the empirical results show that profile learning can reveal which behavioral information continues to improve population-level simulations after rules, incentives, and available actions change. The cross-family comparisons show that the usefulness of behavioral information also depends on the relationship between the game in which it was observed and the game being simulated. Third, because the selected features have substantive definitions and traceable behavioral evidence, their predictive successes and failures can help researchers decide which behavioral regularities to examine in subsequent human experiments, complementing approaches that use prediction and simulation to guide scientific discovery \citep{Gopal2025,AgrawalEtAl2020,LudwigMullainathan2024,ShresthaEtAl2021,ManningZhuHorton2024}.

\section{Background}

\subsection{Language models as behavioral simulators}

Researchers use LLMs to simulate choices and interactions in social and economic settings \citep{Aher2023,Akata2025,Ashokkumar2026}. Many studies condition these simulations on descriptions of the people the model is meant to represent, drawing on survey-based backstories, interviews, or behavior from earlier experiments \citep{Argyle2023,ParkEtAl2026,CrossHaberYamins2025,ManningHorton2025}. Comparisons with human data show that these simulations can reproduce some behavioral patterns, although their accuracy varies across tasks, populations, and design choices \citep{Aher2023,Ashokkumar2026,CrossHaberYamins2025, shah2025learning, ManningHorton2025}.

The effect of a persona depends both on what it contains and on how the model is asked to answer. Individual persona attributes yield modest average gains, and they help most when they explain variation in the human responses being simulated \citep{HuCollier2024}---a condition rarely checked before a persona is written. Across a broad collection of behavioral experiments, \citet{BinzEtAl2026} find little improvement in predictions of individual responses from adding participants' demographic and questionnaire information to prompts. Wording matters as well: simulated responses shift with how a demographic persona is phrased, and personas generated by a model introduce their own systematic deviations \citep{LutzEtAl2025,LiEtAl2025Persona}. A behavioral profile is therefore part of the simulation design, and its value has to be established for the information it carries, the way the model is asked to use that information, and the setting in which the simulation is evaluated.

A complementary line of work designs persona generators to broaden the range of behavior represented in synthetic populations. \citet{PaglieriEtAl2026} define explicit axes of diversity and optimize generators to produce populations spanning those axes, evaluating both diversity in held-out contexts and agreement with human trait distributions. Our framework addresses the related problem of choosing which features of observed behavior help reproduce human outcome distributions. We construct profiles from human experimental records using literature-motivated features, select their contents according to simulation performance, and assess whether those contents remain useful for new participants under new experimental designs.

\subsection{Behavior, context, and predictive generalization}

The practical value of a behavioral profile depends on whether it continues to improve simulation when the setting changes. Behavioral studies typically cover only a limited set of conditions, making it difficult to know which observed patterns will remain predictive elsewhere \citep{AlmaatouqEtAl2024,AlmaatouqNa2026,HofmanEtAl2021}. Prior evaluations of LLM-based behavioral simulation often use well-known games with mature experimental literature, including public-goods games, the prisoner's dilemma, the ultimatum game, and the trust game \citep{Aher2023,Akata2025,SreedharEtAl2025,XieEtAl2024}. These studies provide useful evidence about whether LLMs reproduce established behavioral patterns under a particular experimental design, but provide less evidence about whether their performance persists when the rules or incentives change.

Evaluating a new sample under the same experimental design asks whether a behavioral pattern recurs among different participants. Changing the experimental design asks whether it persists when the rules or incentives change. We focus on this second question because it determines whether profiles learned from earlier experiments remain useful for simulating new designs. Profile contents are chosen under one set of rules, incentives, or available actions and evaluated under another. The two sets contain different participants, and no exact design used to choose a profile is reused to evaluate it. Predictive systems whose performance may deteriorate under a change of setting are better evaluated against this second standard \citep{DuanEtAl2026}.

\subsection{From behavioral records to selectable features}

Raw behavioral records are not organized around the concepts that researchers use to describe behavior. A strategic decision, message, or numerical choice may reflect several behavioral tendencies, while the same tendency may appear through different actions across settings. For example, declining a financial offer may reflect distrust, concern for fairness, or risk aversion, while the same risk aversion may appear as choosing a secure job over an uncertain opportunity. Translating these observations into a common representation therefore requires connecting them to clearly defined social  constructs. Research in information systems and computational social science provides methods for defining behavioral concepts and their indicators, recovering outcome-relevant concepts from language, and deriving theory-guided measures from text \citep{LiLarsenAbbasi2020,Larsen2026,LeeGDCM2025,YangLauAbbasi2022}. Recent work applies LLMs to psychological text analysis \citep{Mousavi2026} and examines how prompts can elicit behavior associated with different possible motivations \citep{Xie2025}. Other work calibrates simulated cognitive biases against classic social science experiments \citep{LiuShangJin2026}.

Representing unstructured behavioral records as clearly defined numerical features makes the information both interpretable and controllable. Each numerical feature is linked to the evidence supporting it, allowing researchers to inspect how it was assigned and compare the same feature across different participants. Because features can be included or excluded individually, researchers can measure how each affects simulation when included in the prompt.

Deciding which features belong in a profile therefore becomes a predictive design problem. A feature may clearly describe earlier behavior and still add little when the simulator sees it; another may materially improve the match between simulated and human choices. Predictive feature selection provides the governing principle: evaluate a possible input by whether it improves predictions beyond the observations used to choose it \citep{GuyonElisseeff2003,ShmueliKoppius2011}. Recent work suggests that persona construction can be adapted to the target simulation task: \citet{BuiEtAl2025} learn how often to sample different persona descriptions and example responses so that simulated responses better match those of a target population. \citet{SorensenEtAl2025} construct natural-language value profiles from observed ratings of individuals and evaluate how much these profiles predict the ratings of the same participant. We ask how to build and choose interpretable features from earlier human behavior so that an LLM simulator continues to reproduce the distribution of human choices after the experimental conditions change.

\section{Methods and Data}
\label{sec:design}

\begin{figure}[htbp]
  \centering
  \captionsetup{font={small,stretch=1.0}}
  \includegraphics[width=\textwidth]{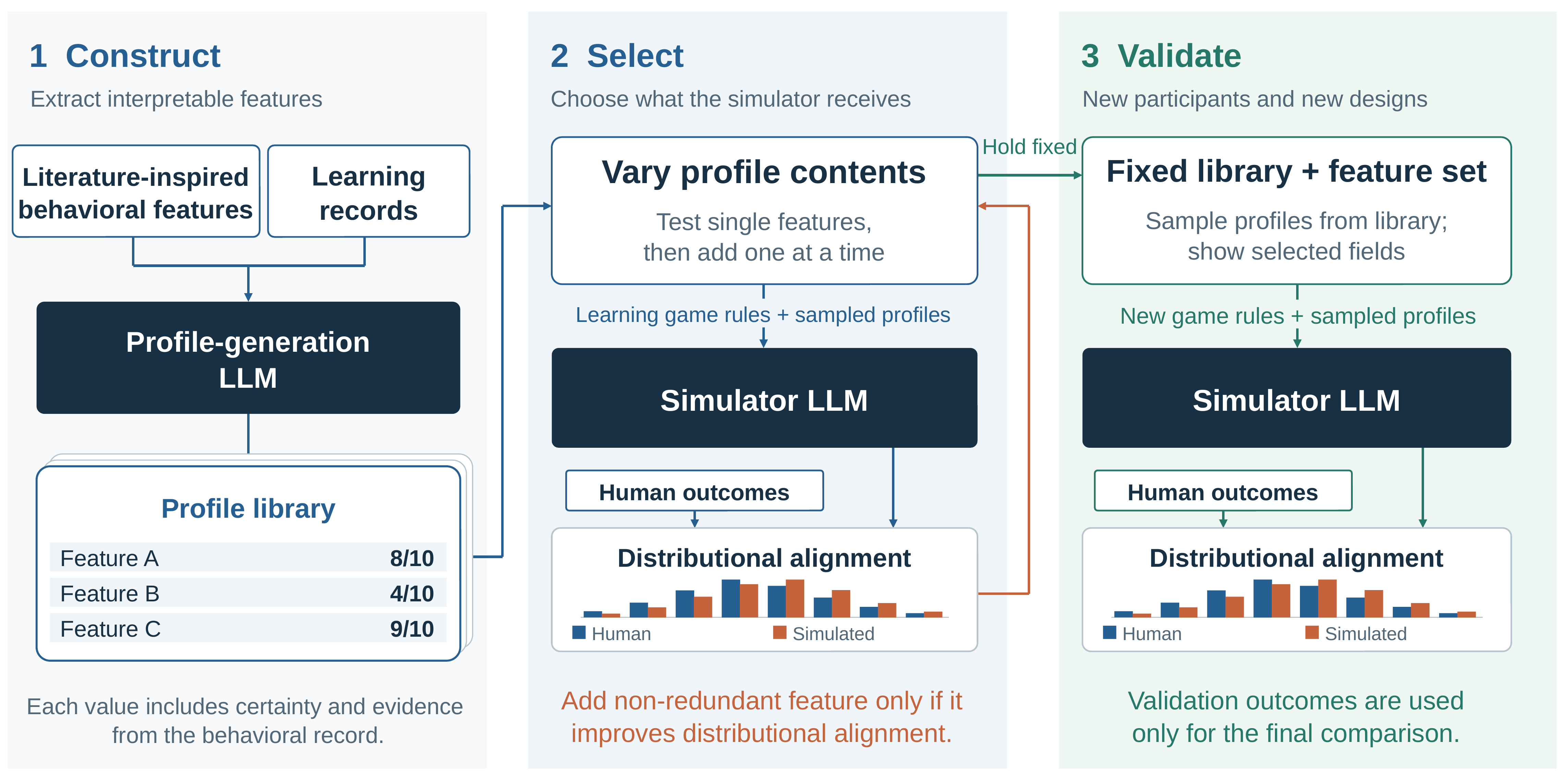}
  \caption{\textbf{From behavioral literature and learning data to validation in new experimental designs.} Construction and selection form the profile learning stage. In each selection comparison, the records used to construct the sampled profiles and the human outcomes used to evaluate simulations come from separate parts of the learning data. Validation tests the chosen profile contents with new participants under new designs. All feature names, scores, and distributions in the schematic are illustrative; empirical results appear in Figure~\ref{fig:heldout}. Candidate behavioral features are defined from the literature and reviewed by authors before any simulation result is examined.}
  \label{fig:design}
\end{figure}

\subsection{Profile learning and validation design}

Figure~\ref{fig:design} shows the three steps of the framework. Profile construction and feature selection form the profile learning stage, followed by validation with new participants under new experimental designs. Within each learning comparison, the profiles used in simulation are drawn from behavioral records separate from the human outcomes against which those simulations are evaluated.

We consider two tests of generalization \citep{hupkes2023taxonomy}. The primary test is \emph{within-family generalization}: the game family stays fixed while the profile is evaluated under new rules, incentives, information, action sets, or interaction structures---when communication, private values, or payoff rules change. The second test is \emph{cross-family transfer}: profiles sampled from a library constructed within one family are used to simulate another, for example, from public-goods experiments to multi-party bargaining. Both tests are evaluated under experimental designs absent from the learning data. Appendix~\ref{oa-sec:generalization} details the data partitions for these comparisons.

\subsection{Three families of incentivized experiments}

We apply the framework to three human-subject experimental studies. Public-goods experiments study repeated decisions about allocating private resources to a shared return \citep{AlsobayEtAl2026}. Multi-party bargaining experiments study three-party exchange when asset values are privately known \citep{QianEtAl2025}. Generalized 11--20 money-request games study strategic numerical choices under varied payoff rules \citep{ManningHorton2025}.

Several properties make these studies well suited to evaluating behavioral profiles. Each has a mature experimental literature from which a broad candidate feature set could be defined before any simulation result was examined. Each also spans a rich space of experimental design such as changing incentive structures, allowing learning and validation to use nonoverlapping designs. Together they span collective resource allocation and peer governance, multilateral exchange under private information, and strategic anticipation under changing incentives. Their behavioral records also differ greatly in depth---from repeated group interaction and multi-party negotiation to a single numerical choice---which tests whether one framework can use evidence of very different richness and social context.

Public versions of all three studies were posted after the reported knowledge cutoffs of the models used in the primary profile-generation and simulation analyses, reducing the risk that the models reproduced the focal results from memory while still allowing them to draw on the older behavioral literature \citep{LudwigMullainathanRambachan2025,AlsobayEtAl2026,ManningHorton2025,QianEtAl2025,OpenAIGPT5Mini,OpenAIGPT51}. Appendix~\ref{oa-subsec:chronology} reports the exact release and cutoff dates.

\subsubsection{Public-goods games}

In each round of repeated public-goods games, participants decide how much of a private endowment to place in a shared account that benefits the group \citep{AlsobayEtAl2026}. The study varies 13 contextual design parameters---among them group size, game length, the return to the shared account, and whether participants can communicate with, reward, or punish one another---and assigns punishment availability separately as its focal manipulation, for 14 manipulated parameters in total, which \citet{AlsobayEtAl2026} report in full. These variations change both the economic return to cooperation and the ways participants can coordinate or govern one another, and they expose behavioral evidence about contributions, responses to peers, communication, and the use of social incentives.

The learning data comprise records from 366 completed games, each with a distinct experimental design, from which we construct 3,691 participant profiles. A panel of 48 games balanced across communication, reward, and punishment is used to compare profile contents, while the remaining 318 provide the profiles available for sampling. The validation data comprise records from 250 completed games across 40 nonoverlapping experimental designs. After feature selection, profiles from all 366 learning games are available for these validation simulations. Profiles are sampled only from learning games with the same availability of communication, reward, and punishment as the game being simulated (Appendix~\ref{oa-sec:source-units}).

Following \citet{AlsobayEtAl2026}, the focal outcome we use to compare between LLM-simulated distribution and human distribution is normalized efficiency: realized group earnings relative to reference earnings under universal non-contribution and universal contribution, with no rewards or punishments in either benchmark. Zero and one correspond to those two reference behaviors; actual rewards can raise efficiency above one, and punishments can lower it below zero.

\subsubsection{Multi-party bargaining}

In the multi-party bargaining study, participants exchange several types of chips whose point values are private and differ across players \citep{QianEtAl2025}. Each group of three participants, which we call a cohort, plays two games under different private-value assignments, and each participant can make and respond to proposals. The setting captures a multilateral allocation problem in which parties must discover mutually beneficial exchanges while negotiating who receives the surplus \citep{zhu2026choose}.

We construct one profile for each of 216 participants from both game records and a post-session participant survey. Profile learning uses data from designs with two or three distinct chip types (48 cohorts, 144 participants, 96 games), whereas validation uses data from designs with four chip types (24 cohorts, 72 participants, 48 games). Simulations sample profiles from the 144 learning participants. We keep both games and all three participants from a cohort together throughout the analysis. Following \citet{QianEtAl2025}, the outcome for each game is the final surplus ratio: the gains from trade realized by the final allocation divided by the maximum attainable from the same initial allocation and private values.

\subsubsection{Generalized 11--20 money-request games}

The classic 11--20 game asks two players to request an integer from 11 to 20. Each receives the requested amount, plus a bonus for requesting exactly one less than the other player, which makes it a standard test of bounded strategic reasoning: the best request depends on both the payoff rule and what a participant expects the other to choose \citep{AradRubinstein2012,CamererHoChong2004}. The generalized study varies six components of this payoff structure, among them the permitted request range, the bonus size, and the relation between requests that earns the bonus, across 11 bonus rules \citep{ManningHorton2025}; together these variations change which requests are permitted, the incentives, and the strategic demands of the task.

Almost every participant makes one request in one design, so a single observed choice cannot support a person-level profile. We instead generate one response-type profile for each unique payoff-design--choice pair and preserve the number of people it represents as its sampling weight. The learning data cover 596 payoff designs, 1,670 human choices, and 1,337 unique response types; the validation data cover 884 unseen designs and 2,502 human choices. To compare requests across different permitted request ranges, we normalize choice $x$ as $(x-L)/(U-L)$ for the design's lower and upper bounds $L$ and $U$, so that zero denotes the smallest permitted request and one the largest.

Table~\ref{tab:study-overview} summarizes the three applications after these study-specific units and outcomes have been defined.

\begin{table}[htbp]
  \centering
  \small
  \caption{\textbf{Three applications of the framework.} The design columns show which experimental designs were used during profile learning and which nonoverlapping designs were reserved for validation with different participants.}
  \label{tab:study-overview}
  \begin{spacing}{1.0}
  \setlength{\tabcolsep}{4pt}
  \begin{tabular}{P{0.12\textwidth}P{0.19\textwidth}P{0.21\textwidth}P{0.22\textwidth}P{0.14\textwidth}}
    \toprule
    Game family & Profile represents & Designs for profile learning & Validation designs & Outcome \\
    \midrule
    Public goods & Participant in one game & 366 game designs & 40 nonoverlapping designs & Normalized efficiency \\
    Multi-party bargaining & Participant across two games and survey & Designs with two or three chip types & Designs with four chip types & Final surplus ratio \\
    Generalized 11--20 & Weighted payoff-design--choice type & 596 payoff designs & 884 nonoverlapping designs & Normalized request \\
    \bottomrule
  \end{tabular}
  \end{spacing}
\end{table}

\subsection{Step 1: Construct behavioral profiles}
\subsubsection{Defining candidate features and scoring rules}

Profile construction begins by specifying which aspects of behavior the profile generator may represent. For each family we surveyed the relevant behavioral and experimental literature and defined a candidate feature set. LLMs assisted with literature synthesis and consistency checks, while the authors reviewed the sources, definitions, and boundaries before simulation outcomes were analyzed. We then developed a \emph{feature codebook} that defines what low and high values mean, anchors low, midpoint, and high values at 0, 5, and 10, states when the setting gives a participant an opportunity to display the behavior, and distinguishes qualifying from nonqualifying evidence. These specifications turn concepts such as withdrawal after adverse peer behavior or sensitivity to payoff incentives into instructions applicable to heterogeneous records. Appendix Tables~\ref{oa-tab:si-codebook-pgg}--\ref{oa-tab:si-codebook-bounds1120} list the candidate features, definitions and literature sources; the accompanying feature-codebook file provides the full numerical anchors and evidence rules. The candidate sets contain 30 public-goods features, 20 bargaining features, and seven generalized 11--20 features.

\subsubsection{Extracting evidence-grounded behavioral profiles}

The profile-generation LLM receives an earlier behavioral record together with the corresponding feature codebook. Depending on the family, the record contains a participant's actions and messages in one repeated game, both bargaining games and the participant survey, or one observed response to a money-request design. For every feature the model returns either a value from 0 to 10 or a status indicating that the record does not support one. A feature is \emph{identifiable} from a behavioral record when the participant had the relevant opportunity to display the behavior and the record provides qualifying direct evidence for a numerical assignment. For every identifiable feature, the profile reports the evidence and where it came from, together with the profile-generation model's certainty in its interpretation. Features without qualifying evidence are left out of the profile.

An identifiable feature appears in the profile in roughly this form:
\begin{quote}
\small\ttfamily
withdrawal after adverse peer behavior: 8/10; certainty 7/10\\
evidence (own contributions): 20, 20, 18, then 4 after two peers gave 0
\end{quote}
The value and evidence shown here are illustrative. Appendix~\ref{oa-sec:profile-generation} describes the prompts and review procedure, and Appendix~\ref{oa-subsec:worked-example} traces an actual behavioral record through feature identification to the profile received by the simulator.

GPT-5.6-Sol separately reviewed suspicious or ambiguous assignments against the feature definitions and behavioral evidence, without access to simulation outcomes or performance measures. Appendix~\ref{oa-sec:features} distinguishes the descriptive frequency of numerical values from eligibility for feature selection; the accompanying identification table reports their frequency and distribution for all candidates.

\subsection{Step 2: Select profile contents}

The profiles constructed from these records form a \emph{profile library} representing the population observed in the learning data. Each sampled profile lists the name, value, certainty, and direct evidence for every identifiable feature. Simulation conditions vary which features are shown, with each feature's name, value, certainty, and supporting evidence included. A one-feature public-goods condition, for example, shows only \emph{withdrawal after adverse peer behavior}, whereas the complete-profile condition shows every feature identifiable from the same behavioral record. Within each comparison, the experimental design, sampled profile, simulator, and model settings remain unchanged. The no-profile condition receives the same game instructions and state without a behavioral profile.

Each simulated participant or decision receives the same sampled profile across profile conditions, so only the displayed features vary. In bargaining, the same sampled profile stays with a participant across both paired games. In generalized 11--20, profiles are sampled within the same bonus-rule family and weighted to preserve human response frequencies. Appendix~\ref{oa-sec:simulation} describes the model settings and information included in the simulator's prompt, and Appendix~\ref{oa-sec:source-units} reports the assignment rules.

We choose profile contents through forward feature selection, which builds a feature set by evaluating one addition at a time \citep{GuyonElisseeff2003}. We first evaluate every eligible feature on its own by comparing simulated outcomes with human outcomes in the learning data. Among features that improve on no profile in every prespecified learning comparison, the feature with the lowest error calculated across the combined learning data becomes the leading feature.

Starting from the leading one-feature profile, we test each remaining eligible feature as an addition. Our stopping rule requires the expanded profile to improve on the current profile in every prespecified learning comparison and across the combined learning data. Let $A_{k-1}$ be the current profile, $\mathcal{E}_k$ the features available at step $k$, and $L_f(A)$ the study-specific simulation error in learning comparison $f$, with $\overline L(A)$ denoting the error recomputed from the combined learning outcomes. Feature $d$ may be added only if
\[
\mathcal{C}_k=\left\{d\in\mathcal{E}_k\setminus A_{k-1}:
L_f(A_{k-1}\cup\{d\})<L_f(A_{k-1})\ \text{for every }f,\ \
\overline L(A_{k-1}\cup\{d\})<\overline L(A_{k-1})\right\},
\]
Among the candidates that satisfy this rule, we add the one with the lowest error across the combined learning data. The search then continues from the expanded profile and stops when no tested addition satisfies the rule.

When the candidate set is large, we make this search more efficient by grouping features with similar value patterns in the learning profiles. Two features are linked when the absolute Spearman correlation between their values is at least 0.70 ($|\rho|\geq0.70$). Features in the same group are treated as near-substitutes, so forward feature selection considers one representative from each group. Features that are not highly correlated with another feature remain separate candidates. Appendix~\ref{oa-sec:feature-selection} gives the identification thresholds and selection rule; Figure~\ref{oa-fig:si-redundancy-correlations} shows the correlation groups used by the redundancy constraint.

We carry both the leading one-feature profile and the two-feature profile retained at the end of forward feature selection into validation, alongside the complete-profile and no-profile conditions.

\subsection{Step 3: Validate under new experimental designs}

The validation target is the distribution of behavior in a new population facing the new designs, compared with outcomes from a distinct group of human participants. We compare full distributions because two populations with the same average can still differ in how variable, asymmetric, or clustered their choices are, which is why related evaluations also compare full response distributions instead of averages alone \citep{Argyle2023,Bisbee2024,MeisterEtAl2025,GordonEtAl2026}.

We compare the behavioral profiles from our method with three general-purpose persona libraries from earlier studies: a persona from Nemotron-Personas-USA \citep{NVIDIANemotron2025}, a natural-language persona drawn directly from Twin-2K-500 \citep{ToubiaEtAl2025}, and an American National Election Studies (ANES) socio-demographic backstory constructed following Argyle et al. \citep{Argyle2023}. These comparisons test whether information selected from behavioral records outperforms general-purpose descriptions drawn from other populations. Appendix~\ref{oa-sec:external-profiles} documents how the personas were sampled and assigned.

Within each study, we group outcomes by the experimental rules: exact designs in public goods, original and alternate private-value games in bargaining, and the 11 bonus-rule families in generalized 11--20. We calculate the Wasserstein distance between simulated and human outcomes within each group and then average these distances. Letting $g\in\mathcal{G}$ index these groups, distributional mismatch is
\[
D(X,H)=\frac{1}{|\mathcal{G}|}\sum_{g\in\mathcal{G}}
W_1\!\left(\widehat F_{Xg},\widehat F_{Hg}\right),
\]
where $W_1$ is the one-dimensional Wasserstein distance, $\widehat F_{Xg}$ is the simulated outcome distribution, and $\widehat F_{Hg}$ is the human distribution. To measure how large this distance is compared to distances between human-subject experiments within the same design, we report a normalized distributional-alignment score:
\[
S(X)=100\,\frac{D_{\mathrm{reference}}-D(X,H)}
{D_{\mathrm{reference}}-D_{\mathrm{human}}}.
\]
Here $D_{\mathrm{reference}}$ is the mismatch produced by the study's uninformed reference, and $D_{\mathrm{human}}$ is the expected mismatch between two human datasets resampled within the prespecified comparison groups and at the observed sample sizes. A score of 0 matches the study's uninformed reference: expected normalized efficiency under uniform contribution in public goods, and uniformly random permitted actions in the other two families. A score of 100 means that it is about as close to the observed outcomes as another human sample of the same size would typically be. Values are not capped, so a condition can fall below the uninformed reference or exceed the estimated human benchmark.

For public goods, feature selection and the comparison of feature rankings across stages use game-level absolute error in normalized efficiency. Validation comparisons between profile conditions use the distributional distance underlying the normalized score.

Bootstrap intervals resample games in public goods, entire cohorts in bargaining, and payoff designs in generalized 11--20. Figure~\ref{fig:heldout} reports 95\% bootstrap intervals for each condition's normalized score; direct comparisons in the text pair simulations on the same games, cohorts, or payoff designs. We assess differences between conditions using 95\% bootstrap intervals for the paired differences in distributional loss. Appendix~\ref{oa-sec:outcomes} gives the reference and resampling details, and Table~\ref{oa-tab:si-paired-comparisons} reports the paired comparisons.

\section{Results}

\subsection{Profile learning distinguished useful from unhelpful features}

Candidate features drawn from the behavioral literature differed sharply in how much they improved the simulations. In every game family, forward feature selection identified a leading feature and then added one other feature. Selection stopped at two because no tested third-feature addition improved every learning comparison and the combined result. Table~\ref{tab:learning-features} summarizes the features chosen at each step and the feature that produced the largest error when evaluated on its own.

\begin{table}[htbp]
\centering
\small
\setstretch{1.0}
\renewcommand{\arraystretch}{1.18}
\caption{\textbf{Features selected and ranked lowest during profile learning.} Citations beside each feature identify representative literature used to define it; the task-specific definitions and scoring rules are ours. Instruction comprehension is a methodological control tied to the source study's rule-comprehension checks, rather than a literature-derived personality construct. The lowest-ranked column reports the one-feature profile with the largest learning-stage error.}
\label{tab:learning-features}
\begin{tabular}{P{0.12\textwidth} P{0.25\textwidth} P{0.25\textwidth} P{0.25\textwidth}}
\toprule
Game family & Leading feature & Second selected feature & Lowest-ranked feature \\
\midrule
Public-goods games & \emph{Withdrawal after adverse peer behavior} \citep{Fischbacher2001,KeserVanWinden2000}\newline Reduces later contributions after peer free riding, conflict, or another adverse social event. & \emph{Rule comprehension} \citep{HouserKurzban2002}\newline Applies the game's payoff and mechanism rules accurately. & \emph{Commitment making} \citep{KerrKaufmanGilliland1994}\newline Makes or elicits explicit, nonbinding promises about future contributions. \\
\addlinespace
Multi-party bargaining & \emph{Package breadth} \citep{QianEtAl2025,ChakrabortyHarbaugh2003}\newline Negotiates larger multi-chip packages rather than minimum-size exchanges. & \emph{Integrative trade search} \citep{ChakrabortyHarbaugh2003}\newline Searches for packages that can create gains for both sides. & \emph{Profitable-offer uptake} \citep{Guth1982}\newline Accepts offers that clearly increase the participant's induced value. \\
\addlinespace
Generalized 11--20 money-request games & \emph{Payoff and incentive sensitivity} \citep{McKelveyPalfrey1995,ManningHorton2025}\newline Responds to the payoff schedule and its material tradeoffs. & \emph{Instruction comprehension and adherence} \citep{ManningHorton2025}\newline Chooses consistently with the stated rules and instruction checks. & \emph{Opponent-belief sophistication and hedging} \citep{AradRubinstein2012,GoereeLouisZhang2018}\newline Adjusts the request to uncertainty about the other player's reasoning. \\
\bottomrule
\end{tabular}
\end{table}

In public-goods games, withdrawal after adverse peer behavior was the leading feature. It measures how sharply a participant reduces later contributions after peer free riding, conflict, or another adverse social event---a response closely connected to the trajectory of group efficiency. With this feature already in the profile, forward feature selection next added rule comprehension, which records whether a participant accurately applied the game's payoff and mechanism rules. At the other end of the ranking, commitment making produced the largest learning error; reward propensity and responsive or affiliative communication also increased error relative to no profile in both learning and validation.

In multi-party bargaining, the simulations benefited most from information about how participants constructed trades. The leading feature, package breadth, describes a participant's tendency to propose larger multi-chip packages. Forward feature selection next added integrative trade search, which describes efforts to assemble packages that could create gains for both sides. Profitable-offer uptake produced the largest learning error; own-surplus acceptance thresholds also increased error in both stages, while endgame urgency helped during learning but not validation.

In generalized 11--20 money-request games, payoff and incentive sensitivity was the leading feature. It describes how closely an observed request follows the incentives of its payoff design. Forward feature selection next added instruction comprehension and adherence, which records whether a request is consistent with the stated rules and instruction checks. Opponent-belief sophistication and hedging produced the largest learning error, although its validation point estimate was slightly positive; indeed, every candidate feature improved on no profile in validation. Across the three game families, theoretical relevance alone did not determine whether a feature would improve simulation.

Appendix Table~\ref{oa-tab:rank-comparisons} reports every eligible one-feature comparison. Table~\ref{oa-tab:si-learning-comparisons} reports the second-feature comparisons and third-feature stopping tests in all three game families.

\subsection{Leading profiles improved validation alignment}

\begin{figure}[t]
  \centering
  \includegraphics[width=\textwidth]{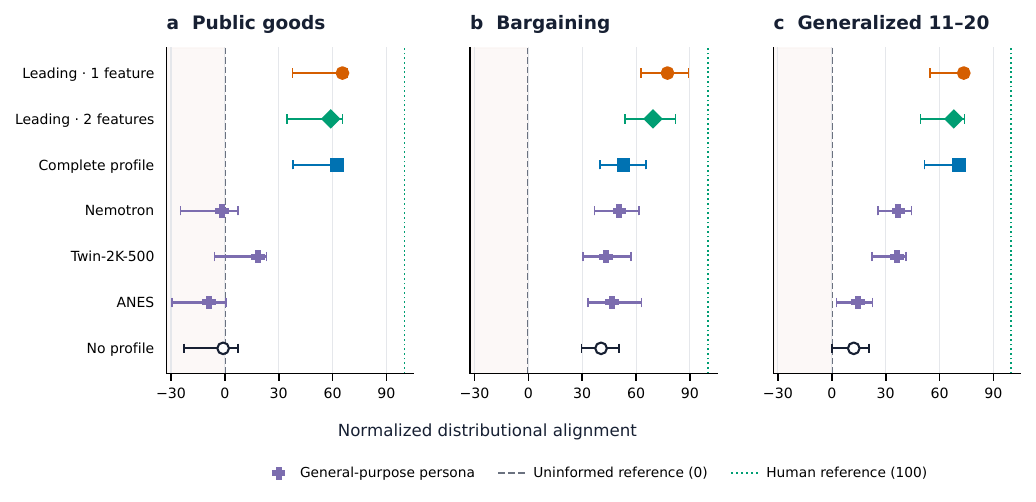}
  \caption{\textbf{Alignment between simulated and human outcomes in new experimental designs.} Each panel presents the same seven comparisons: the leading one- and two-feature profiles, the complete set of identifiable features, three general-purpose persona libraries, and no profile. The general-purpose personas come from Nemotron-Personas-USA records, Twin-2K-500 survey summaries, and ANES socio-demographic backstories following Argyle et al. \citep{Argyle2023,ToubiaEtAl2025,NVIDIANemotron2025}; they are descriptions drawn from different populations, not alternative descriptions of the same people. Points show normalized distributional alignment, and higher values indicate closer agreement with human outcomes. Zero is each study's prespecified uninformed reference and 100 the expected match between two resampled human datasets; the shaded band left of zero marks agreement worse than the uninformed reference. Horizontal lines are 95\% bootstrap intervals for each condition separately, resampling games in public goods, cohorts in bargaining, and payoff designs in generalized 11--20.}
  \label{fig:heldout}
\end{figure}

The validation stage tests whether features chosen under earlier designs remain useful after rules and incentives change. We carry the leading profiles into validation without modification and compare them with no profile, the complete profile, and general-purpose persona libraries. In every family, the leading one-feature profile achieves the highest normalized distributional-alignment score among the conditions shown in Figure~\ref{fig:heldout}. The figure gives condition-specific 95\% intervals; the text reports paired comparisons on the same games, cohorts, or payoff designs.

\paragraph{Public goods games.}

Across all 250 validation games, the leading one-feature profile, withdrawal after adverse peer behavior, achieves a normalized distributional-alignment score of 65.5 (95\% bootstrap interval: [37.6, 68.1]). On this scale, 0 represents the uninformed reference and 100 the expected agreement between two resampled human datasets. A score of 65.5 therefore closes about two-thirds of the gap between these benchmarks. Without a profile, the score is $-1.0$ (95\% bootstrap interval: [$-22.6$, 7.5]); the general-purpose persona scores are 18.6 for Twin-2K-500, $-1.5$ for Nemotron, and $-8.7$ for ANES.

The leading one-feature profile improves distributional alignment over no profile by 66.5 points (paired 95\% bootstrap interval: [42.3, 79.5]). Paired comparisons also favor it over each general-purpose persona library (Appendix Table~\ref{oa-tab:si-paired-comparisons}). Thus, information about how people respond to adverse peer behavior remains useful after the experimental design changes and improves on the three general-purpose persona libraries tested here.

The complete profile scores 62.5 (95\% bootstrap interval: [37.9, 65.2]), and the leading two-feature profile scores 59.0 (95\% bootstrap interval: [34.5, 65.5]). Their paired differences from the leading one-feature profile remain unresolved, providing no clear evidence that either larger profile improves the simulation.

\paragraph{Multi-party bargaining.}

Across the four-chip validation designs, the leading one-feature profile, package breadth, scores 77.5 (95\% bootstrap interval: [62.8, 89.2]). The corresponding scores are 40.5 without a profile (95\% bootstrap interval: [29.6, 50.6]), 53.1 for the complete 20-feature profile (95\% bootstrap interval: [39.9, 65.4]), and 69.5 for the leading two-feature profile (95\% bootstrap interval: [53.9, 82.0]). Package breadth therefore remains more informative than either no profile or the complete profile after the number of chip types and private-value structure change.

\paragraph{Generalized 11--20.}

Across 884 new payoff designs, the leading one-feature profile, payoff and incentive sensitivity, scores 73.5 (95\% bootstrap interval: [54.6, 75.9]). The corresponding scores are 12.2 without a profile (95\% bootstrap interval: [0.2, 20.7]), 70.8 for the complete seven-feature profile (95\% bootstrap interval: [51.6, 74.0]), and 68.0 for the leading two-feature profile (95\% bootstrap interval: [49.5, 73.8]). Payoff sensitivity therefore remains the leading description when the permitted request ranges, bonus rules, and payoff schedules change.

This result applies to the request distribution pooled across designs within groups that share the same bonus rule. A simulation can reproduce the overall frequency of different requests while predicting them under the wrong payoff designs. Secondary comparisons instead score each simulated request against human choices for that particular payoff design; they favor the no-profile condition over the leading and complete profiles. The aggregate improvement therefore does not establish better prediction of responses to a particular change in payoff rules. Appendix~\ref{oa-subsec:1120-diagnostics} reports the metrics and results.

\subsubsection{Leading profiles and general-purpose personas}

General-purpose personas sometimes improve on no profile, but their effects vary across games and none matches the leading one-feature profile's point estimate. The strongest scores 18.6 in public goods games, 50.8 in multi-party bargaining, and 37.1 in generalized 11--20. Reweighting these persona libraries using the learning data does not close the gap in any family (Appendix~\ref{oa-subsec:external-calibration}).

\subsection{Earlier experiments anticipated which features would transfer}

\begin{figure}[htbp]
  \centering
  \includegraphics[width=\textwidth]{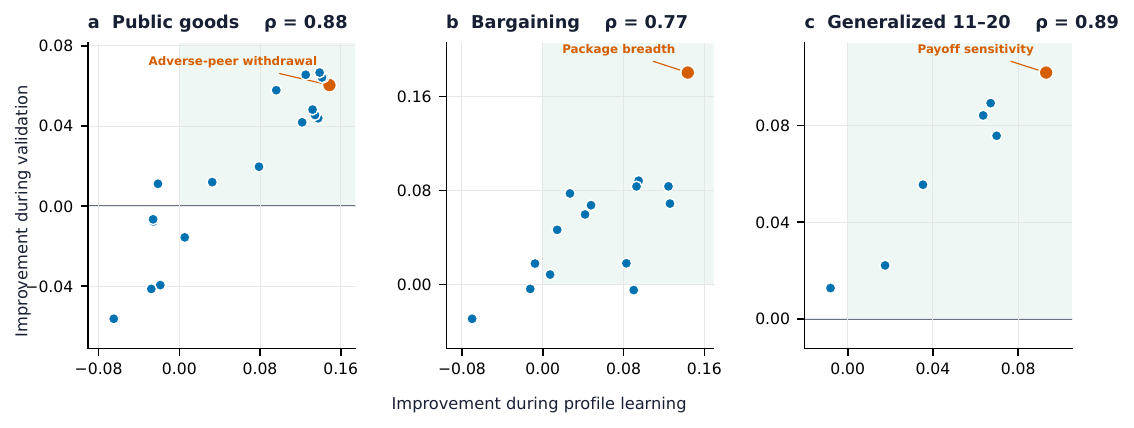}
  \caption{\textbf{Transfer of one-feature usefulness from profile learning to validation.} Each point is one feature evaluated in both stages. Both axes report the reduction in error relative to no profile, so positive values indicate improvement and the origin marks the no-profile condition. \textbf{a,} Public goods uses the reduction in game-level absolute error in normalized efficiency ($\rho=0.884$ across 19 features); its validation values use the same 250 games as Figures~\ref{fig:heldout} and \ref{fig:cross-family}. \textbf{b,} Bargaining uses the reduction in one-dimensional Wasserstein distance between final-surplus-ratio distributions, averaged across experimental designs ($\rho=0.768$ across 15 features). \textbf{c,} Generalized 11--20 uses the reduction in Wasserstein distance between normalized-request distributions, giving equal weight to the 11 groups of games that share the same bonus rule ($\rho=0.893$ across seven features).}
  \label{fig:transfer}
\end{figure}

The relative predictive value of individual features also persists from learning to validation in all three families (Figure~\ref{fig:transfer}). The ordering in public goods has a Spearman rank correlation of 0.884 across 19 features, and four of the five leading features in learning remain in the top five in validation. The bargaining ordering has a correlation of 0.768 across 15 features when moving from designs with two or three chip types to designs with four chip types. Generalized 11--20 has a correlation of 0.893 across seven features, and payoff sensitivity remains the leading individual feature. The broad agreement between these rankings shows that the learning data contained information about which behavioral descriptions would remain useful after the experimental design changed.

\subsection{Profiles transferred unevenly across games}
\label{sec:crossfamily}

Cross-family transfer tests how far this portability extends by using a profile built and selected for one game family to simulate another while leaving the target games and simulator unchanged (Appendix~\ref{oa-sec:cross-family}).

\begin{figure}[htbp]
  \centering
  \includegraphics[width=\textwidth]{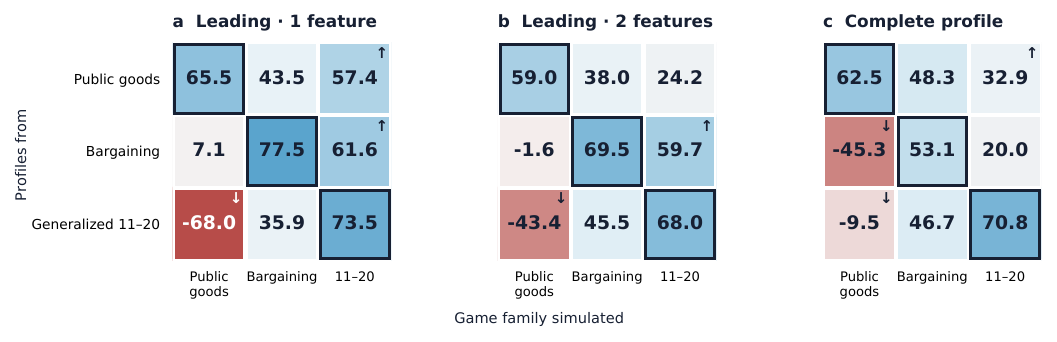}
  \caption{\textbf{Behavioral profiles predict best within the game family used to construct them and transfer asymmetrically.} Rows show the game family used to construct the profile; columns show the family being simulated. Panels use the leading one-feature profile, the leading two-feature profile, or every feature identifiable from the behavioral records. Cells report normalized distributional alignment on the same scale as Figure~\ref{fig:heldout}; blue shading marks positive alignment and red shading marks scores below the uninformed reference. Dark outlines mark within-family predictions, which are highest in every target column. Off-diagonal values differ sharply by direction and several fall below the uninformed reference, so transfer across families is both incomplete and asymmetric. Arrows mark a clear improvement ($\uparrow$) or decline ($\downarrow$) relative to simulation without a profile in the target family, meaning that the 95\% bootstrap interval for the difference excludes zero; no arrow indicates an inconclusive difference.}
  \label{fig:cross-family}
\end{figure}

The profile selected inside a family gives the closest match to that family's human outcomes in all three cases---65.5 in public goods, 77.5 in bargaining, and 73.5 in generalized 11--20 for the one-feature profiles. No profile constructed for another family performs as well. The effect of a profile from another family depends on the direction of transfer. Bargaining and public-goods profiles both improve simulation of generalized 11--20. The bargaining profile also raises the public-goods alignment score, but its paired difference from no profile is inconclusive. The generalized 11--20 profile applied to public goods scores $-68.0$, far below the uninformed reference and clearly worse than simulation without a profile, while neither transfer into bargaining differs clearly from no profile. Adding features does not repair this and in places weakens it: the two-feature generalized 11--20 profile remains clearly harmful in public goods at $-43.4$, and the complete profiles from both other families fall below the uninformed reference when applied to public goods. The usefulness of a profile therefore depends on the relationship between the behavior it describes and the interaction being simulated. A profile that improves simulation within one game family can reduce accuracy when applied to another.

\begin{samepage}
Compact profiles also have lower estimated error than no profile across alternative profile-generation and simulation models, although the selected features change (Appendix~\ref{oa-sec:model-roles}).
\par
\end{samepage}

\section{Discussion}

When generative models are used to study human behavior, the descriptions in their prompts help determine the evidence on which researchers will base their conclusions. Grounding those descriptions in observed behavior allows researchers to inspect how they were constructed, but leaves open how an LLM will interpret them and whether that interpretation will result in useful predictions and simulations. The distinction matters because a description can be supported by the record and still make a simulation less accurate. Our results show that the predictive value of such descriptions can be assessed in earlier experiments and can persist when experimental designs change.

\subsection{What profile learning reveals}

Features drawn from the same literature differed substantially in their value to the simulator. In public goods, withdrawal after adverse peer behavior reduced error in both learning and validation, while commitment making and reward propensity increased it in both. In bargaining, profitable-offer uptake and own-surplus acceptance thresholds also increased error in both stages. The profile learning stage allows researchers to choose among these descriptions by testing their usefulness for the behavior being simulated.

The leading profiles improved simulations of different participants facing rules, incentives, and available actions that did not appear during selection. The agreement between learning and validation rankings extends this result beyond the leading feature: the earlier data anticipate which descriptions will be most useful under new designs. This generalization depended on the setting. Using profiles constructed for another game family improved simulation in some directions, produced inconclusive differences in others, and sometimes increased error. The within-family results therefore provide a basis for using profiles under new experimental designs, while the cross-family results show why a broader application requires its own evaluation.\par The direction of transfer may also reflect differences in the behavioral records used to construct profiles. Leading one-feature profiles drawn from repeated public-goods interactions and bargaining records improved aggregate alignment in generalized 11--20, whereas profiles inferred from individual money requests worsened public-goods simulation and provided no clear benefit in bargaining. Richer records may reveal behavioral tendencies that remain useful across a wider range of settings, while sparser records do not provide enough candidate behavioral features that can be inferred.

General-purpose personas sometimes improved simulation, but their alignment-score point estimates remained below those of the leading one-feature profiles in all three families. These comparisons suggest that profile contents should be chosen by testing how well they help simulations reproduce the human behavior of interest. Our framework combines candidate features drawn from prior research, profile generation tied to behavioral evidence, and selection using learning data. The general-purpose persona libraries differ in the populations, formats, and information they contain, so the comparisons do not isolate which differences account for the performance gap.

\subsection{Why compact profiles may suffice}

The leading one-feature profile outperformed the complete profile in bargaining, with a 95\% bootstrap interval for the difference in error that excluded zero. The corresponding intervals in public goods and generalized 11--20 included zero, leaving these differences inconclusive. Although adding a second feature improved performance during learning in every family, it provided no clear additional benefit in validation. The relative usefulness of individual features largely persisted under new designs, while the benefit of combining them was less consistent.

Compactness here describes what a particular simulator can use, not the dimensionality of the people being described. When we changed the models used to generate and simulate profiles, the leading profiles remained compact, but the features they contained changed. A feature's usefulness therefore depends on how it is inferred from the record and how the simulator responds to it. The selected features provide interpretable descriptions that help a particular simulation. Their predictive value does not establish that they measure stable psychological traits.

Additional features may give the simulator redundant or conflicting guidance about how to act. Studies of persona construction also suggest that adding detail can reduce variation among simulated agents \citep{BhattacharyyaEtAl2026}. Requiring each addition to improve simulation across the learning comparisons may also exclude information whose apparent usefulness is confined to particular experiments. The present comparisons do not distinguish these explanations.

\subsection{Building representations for generative systems}

The study treats the information included in a generative system's prompts as part of the research design. In comparisons among behavioral profiles constructed from the same behavioral records, we held the experimental designs, assigned profiles, and model settings fixed while varying the features shown to the simulator. This allowed us to assess how the choice of profile contents affected simulation. Generic demographic text and length-matched generic text also shifted simulated behavior (Appendix~\ref{oa-subsec:generic-controls}). Comparing those shifts with human outcomes is necessary to establish whether the added information helps.

Three choices make the profile contents and their evidentiary basis traceable. Each feature value is tied to the observation on which it rests; the model's certainty is reported beside the value instead of being folded into it; and a feature the record cannot support is left out, so an absent opportunity never reaches the simulator as a neutral score. These records allow researchers to inspect how the model translated observed behavior into a description. Machine-generated variables can carry error into the analyses that consume them \citep{SchecterLi2025,Zhang2026}. Evaluating the profiles in simulation measures the predictive consequences of using those descriptions, although it does not by itself reveal whether a failure arose in profile generation or in the simulator's use of the profile.

Using an LLM to interpret behavioral records makes it feasible to apply common feature definitions within each game family to thousands of records containing different actions and forms of communication. The LLM simulator then uses those profiles to generate choices and interactions under new rules. Researchers remain responsible for defining the features, deciding what counts as evidence, separating learning from validation, and choosing the outcomes against which simulations are evaluated.

\subsection{Scope, limitations, and future work}
\label{sec:limits}

\paragraph{The representation.} The human experiments provide neither independent benchmark measures nor human-coded ratings for most candidate features. Our quality checks assess consistency with the stated definitions and whether each assignment cites qualifying evidence; they do not establish construct validity. A simulation can reproduce an outcome distribution well even if the profile misinterprets the behavior in the record used to construct it. Each displayed feature includes its name, value, certainty, evidence, and added prompt text, so the design estimates the value of including that entire description in the simulator's prompt. We also do not compare the profiles with prompts containing the complete behavioral records, which could preserve predictive detail lost during profile generation. The unit represented by a profile differs across studies: one participant in one completed public-goods game, the same person's two games and survey in bargaining, and a weighted game--choice response type in generalized 11--20. The last describes response types across payoff designs rather than stable individual personalities.

\paragraph{The evaluation.} Our evidence concerns population-level outcome distributions. The framework samples profiles representing one population and simulates behavior for a different population facing new experimental designs. It estimates the frequencies and collective consequences of actions without identifying which particular person will take each action; treating synthetic participants as substitutes for humans elsewhere would require stronger assumptions and additional validation \citep{HullmanEtAl2026}.

Our three empirical applications do not examine differences across demographic groups. Studies of LLM-generated personas and opinions have documented demographic stereotyping \citep{ChengEtAl2023,LiuEtAl2024} and distorted differences between groups \citep{Bisbee2024}. Whether constructing and selecting behavioral profiles introduces or amplifies such biases therefore remains an open question. Future studies should examine this possibility in other empirical settings by assessing both the profiles' descriptions of different demographic groups and how accurately the simulations reproduce each group's behavior.

\paragraph{The implementation.} The leading features may change with the models and response protocols used to create and use a profile: a different generator may infer different values, a different simulator may use the same value differently, and an output protocol requesting probabilities rather than one action may change how population variation is expressed \citep{GordonEtAl2026}. Candidate sets are bounded by the literature used to build them, and forward feature selection is not exhaustive over subsets. It can miss features that become useful only in combination, and its redundancy constraint may choose one representative even when a correlated alternative would work better in combination.

Several comparisons would clarify these limits. Independent ratings and established instruments could test whether the inferred feature values serve as psychological measurements. Comparing leading profiles, broader structured profiles, and complete behavioral records under one simulator and response protocol would show whether profile generation discards useful information. Future research and managerial applications could introduce an exploration probability, analogous to $\varepsilon$ in reinforcement learning, to occasionally explore alternative starting features or feature combinations using only the learning data \citep{SuttonBarto2018}. This could help uncover combinations missed by the forward search, while machine-suggested candidates under human review could broaden the candidate set.

The substantive definitions of the leading features also suggest questions for new human experiments. Researchers could vary when adverse-peer withdrawal, package breadth, or payoff sensitivity should matter and test the resulting predictions directly. The learning and validation results give a reason to investigate these regularities; new human experiments can establish whether they reflect causal mechanisms and where their usefulness ends \citep{AgrawalEtAl2020,LudwigMullainathan2024,ManningZhuHorton2024}.

\ifblind\else
\section*{Acknowledgments}

This research was supported by the Cooperative AI Foundation funding. The authors thank Abdullah Almaatouq, Mohammed Alsobay, John J. Horton, and Benjamin S. Manning for sharing their experimental data and helping us understand and use them. The authors also thank Olivier Toubia and anonymous reviewers at the NeurIPS 2026 Workshop on User Simulation for thoughtful feedback. In preparing this paper, the authors utilized generative AI models as tools to assist with coding, study design, and manuscript editing.

\fi

\bibliographystyle{arxiv_assets/informs2014}
\begingroup
\small
\bibliography{arxiv_assets/references}
\endgroup

\clearpage
\appendix
\setcounter{table}{0}
\setcounter{figure}{0}
\setcounter{equation}{0}
\renewcommand{\thetable}{A\arabic{table}}
\renewcommand{\thefigure}{A\arabic{figure}}
\renewcommand{\theequation}{A\arabic{equation}}
\renewcommand{\theHtable}{appendix.\arabic{table}}
\renewcommand{\theHfigure}{appendix.\arabic{figure}}
\renewcommand{\theHequation}{appendix.\arabic{equation}}
\section*{Appendix}
\addcontentsline{toc}{section}{Appendix}

\FloatBarrier
\section{Data partitions and generalization}
\label{oa-sec:source-units}\label{oa-sec:generalization}

The learning and validation data contain different participants and nonoverlapping experimental designs. Within-family validation changes rules, incentives or available actions while holding the game family fixed. Cross-family comparisons instead change the family from which the profiles were constructed. Both comparisons assess how closely simulated and human outcome distributions agree.

\subsection{Public goods}

The original study contains repeated group-contribution games under combinations of communication, punishment, and reward \citep{AlsobayEtAl2026}. Each profile describes one participant's behavior in a completed game, using only information available to that participant during the game. Wave 1 provides the learning data: 366 completed games from which we construct 3,691 profiles. A fixed panel of 48 games, balanced across communication, punishment, and reward availability, provides the human outcomes used to compare profile contents. Profiles assigned in these simulations are sampled from the remaining 318 games. Wave 2 provides validation data from 250 completed games. Once feature selection is complete, profiles from all 366 learning games are eligible for sampling in validation simulations. In both stages, profiles are sampled from Wave 1 games with the same availability of communication, punishment, and reward as the game being simulated.

\subsection{Multi-party bargaining}

The bargaining study contains 72 groups of three participants, which we call cohorts, each playing two games with different assignments of private chip values \citep{QianEtAl2025}. One person-level profile combines both game records and a post-session participant survey. Both games and all three participants in a cohort remain together in every split. The same sampled profile stays with each simulated participant across the paired games.

The main analysis uses two- and three-chip designs for profile learning and four-chip designs for validation. The learning data contain 48 cohorts, 144 participants, and 96 games; the validation data contain 24 cohorts, 72 participants, and 48 games. Within the learning data, we divide the 48 cohorts into two groups, each containing 12 two-chip and 12 three-chip cohorts. Each learning comparison simulates eight prespecified cohorts from one group using profiles sampled from the full set of 24 cohorts in the other group. We then reverse the groups' roles and use both comparisons to select features. In either comparison, profiles may be sampled from both two- and three-chip designs in the other group. The validation analysis reported here uses all 24 four-chip cohorts and both games from every cohort. The learning comparisons, sampled profiles, and stopping rule are fixed before validation.

\subsection{Generalized 11--20 money-request games}

Each generalized 11--20 payoff design specifies a permitted request range, a rule translating each request into base points, and a bonus rule \citep{ManningHorton2025}. Almost every person makes one decision, so the profile unit is a unique payoff-design--choice response type. If several people make the same choice under one design, the profile is extracted once and receives their frequency as its sampling weight. The learning data cover 596 payoff designs, 1,337 unique response types, and 1,670 observed human choices. The validation data cover 884 different payoff designs and 2,502 human choices included in the analysis. Within the learning data, payoff designs are assigned to two groups, balancing the bonus-rule families and keeping all human responses to a design together. Simulations evaluated against one group's human outcomes draw profiles only from the other group, with profiles sampled from designs in the same bonus-rule family. We repeat the comparison with the groups' roles reversed and use both comparisons to select features.

\subsection{Dataset release and model chronology}
\label{oa-subsec:chronology}

The primary profile-generation and simulation analyses used the dated \texttt{gpt-5-mini-2025-08-07} and \texttt{gpt-5.1-2025-11-13} model snapshots. OpenAI reports knowledge cutoffs of May 31, 2024 for GPT-5-mini and September 30, 2024 for GPT-5.1 \citep{OpenAIGPT5Mini,OpenAIGPT51}. The studies were first posted publicly on August 23, 2025 for public goods, August 24, 2025 for generalized 11--20, and September 11, 2025 for bargaining \citep{AlsobayEtAl2026,ManningHorton2025,QianEtAl2025}. This chronology reduces the risk that the models reproduced the focal datasets or reported results from memory.

\FloatBarrier
\section{Constructing profiles and simulating behavior}
\label{oa-sec:profile-generation}
The profile-generation model receives the behavioral record, the game family's feature definitions and 0/5/10 anchors, opportunity and evidence rules, and a specified response format. It assigns a numerical value only when the record contains both the relevant opportunity and qualifying direct evidence. Each numerical assessment includes certainty from 0 to 10, an indication of whether the participant had an opportunity to display the behavior, and up to three evidence items in public goods and generalized 11--20 or four in bargaining. Features that cannot be identified are omitted from the profile shown to the simulator.

Public-goods profiles use the actions, messages and game information available to the participant during one completed game. Bargaining profiles combine both games and the post-session survey; survey statements cannot override contradictory actions, and the profile generator is instructed not to infer other participants' private chip values. Generalized 11--20 profiles represent a unique payoff-design--choice pair, weighted by its frequency in the human data.

The profile generator is instructed to score every candidate feature using its own anchors and only information available to the participant. It returns either an evidence-supported value or a status indicating why the record does not support a numerical value. Programmed checks verify the response format, feature identifiers and required evidence. GPT-5.6-Sol reviews flagged or ambiguous cases to assess whether the evidence fits the feature definition, concerns the correct participant, follows the sequence of events and payoff rules, and uses only information available to that participant. It receives neither simulation outcomes nor measures of simulation performance. The authors review these cases and decide whether to exclude a feature. In public goods, P30 is excluded because its generated evidence did not reliably distinguish commitments from other statements, leaving 29 features for complete profiles.

\subsection{A behavioral record and its resulting profile}
\label{oa-subsec:worked-example}
An example from the generalized 11--20 learning data illustrates the distinction between an identifiable feature and an unsupported interpretation. A participant chose 6 from the permitted range 1--6. Base points equaled the request, and both players earned one extra point if their requests differed by more than one. Choosing 6 secured the highest base payoff. The generator assigned payoff and incentive sensitivity a value of 9/10 and certainty of 8/10, citing the highest guaranteed payoff and absence of sacrificed base points. It left strategic iteration depth unidentified because the same choice was consistent with several levels of reasoning.

The profile was subsequently assigned in a learning simulation with requests from 17 to 23, base points equal to the request minus two, and a 13-point bonus if requests differed by more than four. The simulator received the new rules and this recorded profile excerpt:
\begin{quote}\small
\textbf{B03 --- Payoff and incentive sensitivity}\\
Value: 9/10; Confidence: 8/10\\
Direct source evidence: is\_maximin = true and min\_payoff = 6. Choice yields highest guaranteed payoff (no points sacrificed).; Choice = 6; points\_sacrificed = 0; is\_br\_uniform = true, indicating alignment with payoff-maximizing incentives under uniform beliefs.\\
Interpretation: The observed choice aligns closely with material payoff incentives (maximin and best-response under uniform belief), indicating strong payoff sensitivity.
\end{quote}
Here ``confidence'' is the recorded field name for certainty. The evidence fields describe the earlier choice: \texttt{is\_maximin} indicates whether it maximizes the minimum possible payoff; \texttt{min\_payoff} gives that minimum; \texttt{points\_sacrificed} records base points forgone relative to the largest permitted request; and \texttt{is\_br\_uniform} indicates whether it maximizes expected payoff when all permitted requests by the other player are assumed equally likely. The simulator was instructed that the earlier game differed and that it should not copy the literal request. Strategic iteration depth remained absent even in the complete-profile condition.

\subsection{Model roles and simulation information}
\label{oa-sec:simulation}
GPT-5-mini-2025-08-07 generates the primary public-goods and generalized 11--20 profiles; GPT-5.1-2025-11-13 generates the primary bargaining profiles. The primary simulator is GPT-5-mini-2025-08-07. The primary profile-generation models were chosen before their validation results were examined. Both stages use the API's ``low'' reasoning-effort setting, given that the internal reasoning traces through OpenAI's API calls are not retrievable. Instead, we ask each LLM request for explanations of making such predictions about simulation. Model comparisons in Appendix~\ref{oa-sec:model-roles} vary the generator or simulator.

The simulator receives the game rules, available actions, timing, payoffs and information visible to its assigned role. A profile condition adds identifiable features with their names, values, certainty and evidence; the no-profile condition omits the behavioral-profile portion of the prompt. Public-goods and bargaining simulations produce complete interactions, whereas generalized 11--20 simulations return a brief rationale and one permitted integer. Bargaining players see their own holdings and private values, not their counterparts' values.

Within a comparison, the game information, model settings and sampled profiles are fixed while the displayed feature set changes. A bargaining profile remains attached to the same simulated participant across both games. Generalized 11--20 profiles are sampled within bonus-rule families with sampling probabilities proportional to the number of human choices each profile represents. The comparison measures the contribution of each displayed feature together with its evidence and interpretation; it does not isolate the numerical value from its accompanying text.

\FloatBarrier
\section{Candidate features and identification}
\label{oa-sec:features}
The literature motivates the behavioral distinctions below; the authors defined the task-specific anchors, evidence requirements and interpretation limits before simulation outcomes were examined. Tables~\ref{oa-tab:si-codebook-pgg}--\ref{oa-tab:si-codebook-bounds1120} list all 57 candidates. The accompanying \texttt{feature\_codebooks.tsv} retains the full 0/5/10 anchors and evidence rules, including the opportunities needed to assign a value.

The numerical-value rate is the share of records containing a number, weighted by the number of human choices each response-type profile represents in generalized 11--20. It is distinct from eligibility for feature selection, which also requires qualifying assessments and relevant opportunities. For example, public-goods P07 has numerical values in 74.6\% of records, but qualifying assessments in only 56.0\% of records with the relevant opportunity, below the 70\% eligibility threshold. The companion \texttt{feature\_identification.tsv} preserves all descriptive rates and value distributions. Appendix~\ref{oa-sec:feature-selection} states the eligibility rules.

\begingroup
\small
\setlength{\tabcolsep}{3pt}
\setlength{\LTcapwidth}{\linewidth}
\begin{longtable}{P{0.055\linewidth}P{0.59\linewidth}P{0.29\linewidth}}
\caption{Candidate behavioral features: Public goods. Literature motivates the constructs; definitions and scoring anchors are task-specific operationalizations.}\label{oa-tab:si-codebook-pgg}\\
\toprule
ID & Feature and definition & Literature \\
\midrule
\endfirsthead
\multicolumn{3}{l}{\textit{Continued from preceding page}}\\
\toprule
ID & Feature and definition & Literature \\
\midrule
\endhead
P01 & \textbf{Baseline cooperation propensity}. Typical contribution level before accounting for responses to other participants' later behavior. & \citet{Ledyard1995}; \citet{Fischbacher2001} \\
P02 & \textbf{Costly contribution orientation}. Willingness to sacrifice own immediate material payoff through contribution when contribution is privately costly. This is a behavioral tradeoff, not a claim that altruism, efficiency, warm glow, or conformity is the cause. & \citet{Ledyard1995}; \citet{Fischbacher2001}; \citet{Andreoni1990}; \citet{CharnessRabin2002} \\
P03 & \textbf{Opportunistic withholding}. Withholding contribution while benefiting from substantial peer contribution, including late or situational exploitation. & \citet{Ledyard1995}; \citet{Fischbacher2001}; \citet{Andreoni1988} \\
P04 & \textbf{Peer-responsive cooperation}. Degree to which contribution changes in the same direction as observed peer cooperation or explicit cooperative coordination. & \citet{Fischbacher2001}; \citet{KeserVanWinden2000} \\
P05 & \textbf{Threshold cooperation}. Sharply changing contribution levels when peer cooperation crosses a salient level. & \citet{Fischbacher2001} \\
P06 & \textbf{Contribution volatility}. Within-game variability in contribution after accounting for clear peer-contingent responses. & \citet{Ledyard1995}; \citet{Chaudhuri2011} \\
P07 & \textbf{Directional learning trend}. Gradual directional change in contribution as rounds and experience accumulate. & \citet{Andreoni1988}; \citet{Chaudhuri2011} \\
P08 & \textbf{Withdrawal after adverse peer behavior}. Reduction in subsequent contribution after peer free riding, conflict, or another clearly adverse social event. & \citet{Fischbacher2001}; \citet{KeserVanWinden2000}; \citet{FalkEtAl2018} \\
P09 & \textbf{Recovery after peer improvement}. Restoration of contribution after peers resume cooperation or repair a negative interaction. & \citet{KeserVanWinden2000}; \citet{FalkEtAl2018} \\
P10 & \textbf{Endgame contribution change}. Direction of contribution change near the known or reasonably anticipated end of the game. & \citet{Andreoni1988}; \citet{Chaudhuri2011} \\
P11 & \textbf{Initial cooperative optimism}. Initial willingness to contribute before observing how much other participants contribute. & \citet{Fischbacher2001}; \citet{FalkEtAl2018} \\
P12 & \textbf{Exploitation sensitivity}. Behavioral sensitivity to being disadvantaged by low-contributing or higher-earning peers. & \citet{FehrSchmidt1999}; \citet{BoltonOckenfels2000}; \citet{CharnessRabin2002} \\
P13 & \textbf{Rule comprehension}. Accuracy and consistency in applying game and payoff rules. & \citet{HouserKurzban2002} \\
P14 & \textbf{Forward planning}. Evidence that current actions anticipate later rounds, future reactions, or consequences of the game rules. & \citet{Andreoni1988}; \citet{Chaudhuri2011} \\
P15 & \textbf{Punishment propensity}. Frequency and intensity of punishment when it is available. & \citet{FehrGachter2000} \\
P16 & \textbf{Low-contributor punishment selectivity}. Degree to which punishment is concentrated on lower contributors rather than applied indiscriminately. & \citet{FehrGachter2000}; \citet{HerrmannEtAl2008} \\
P17 & \textbf{Punishment direction}. Whether punishment is directed toward higher or lower contributors. & \citet{FehrGachter2000}; \citet{HerrmannEtAl2008} \\
P18 & \textbf{Cooperation response after received punishment}. Direction of the next contribution response after the player is punished. & \citet{FehrGachter2000}; \citet{Nikiforakis2008} \\
P19 & \textbf{Reward propensity}. Frequency and intensity of costly reward when it is available. & \citet{AndreoniHarbaughVesterlund2003}; \citet{SeftonEtAl2007} \\
P20 & \textbf{High-contributor reward selectivity}. Degree to which reward is concentrated on higher contributors. & \citet{AndreoniHarbaughVesterlund2003}; \citet{SeftonEtAl2007} \\
P21 & \textbf{Relationship-contingent reward}. Use of reward conditional on a specific partner's prior favorable treatment or reciprocal interaction. & \citet{AndreoniHarbaughVesterlund2003}; \citet{CharnessRabin2002} \\
P22 & \textbf{Cooperation response after received reward}. Direction of the next contribution response after the player receives reward. & \citet{AndreoniHarbaughVesterlund2003}; \citet{SeftonEtAl2007} \\
P23 & \textbf{Chat engagement}. Frequency of participation when a communication channel is available. & \citet{Sally1995}; \citet{Balliet2010}; \citet{BochetEtAl2006} \\
P24 & \textbf{Coordination leadership}. Attempts to initiate, organize, or stabilize a group contribution plan. & \citet{KoukoumelisEtAl2012}; \citet{BochetEtAl2006} \\
P25 & \textbf{Commitment making}. Making or eliciting explicit nonbinding promises about future contribution. & \citet{KerrKaufmanGilliland1994}; \citet{KoesslerEtAl2021} \\
P26 & \textbf{Cooperation-norm appeals}. Invoking fairness, duty, reciprocity, or a cooperation norm to influence the group. & \citet{HaruvyEtAl2017}; \citet{BochetEtAl2006} \\
P27 & \textbf{Responsive and affiliative communication}. Directly responding to others while maintaining a supportive, appreciative, or rapport-building tone. & \citet{HaruvyEtAl2017}; \citet{Balliet2010} \\
P28 & \textbf{Blame and conflict escalation}. Use of accusation, shaming, hostility, or escalating interpersonal conflict. & \citet{HaruvyEtAl2017}; \citet{BochetEtAl2006} \\
P29 & \textbf{Repair communication}. Apologizing, acknowledging error, conceding, or attempting to restore cooperation after a rupture. & \citet{BottomEtAl2002} \\
P30 & \textbf{Message-action inconsistency}. Degree to which claims, promises, or strategic messages conflict with later observable action. & \citet{KerrKaufmanGilliland1994}; \citet{KoesslerEtAl2021}; \citet{BottomEtAl2002} \\
\bottomrule
\end{longtable}
\endgroup

\begingroup
\small
\setlength{\tabcolsep}{3pt}
\setlength{\LTcapwidth}{\linewidth}
\begin{longtable}{P{0.055\linewidth}P{0.59\linewidth}P{0.29\linewidth}}
\caption{Candidate behavioral features: Multi-party bargaining. Literature motivates the constructs; definitions and scoring anchors are task-specific operationalizations.}\label{oa-tab:si-codebook-chip}\\
\toprule
ID & Feature and definition & Literature \\
\midrule
\endfirsthead
\multicolumn{3}{l}{\textit{Continued from preceding page}}\\
\toprule
ID & Feature and definition & Literature \\
\midrule
\endhead
C01 & \textbf{Own-surplus protection}. Consistency with which choices protect or increase the participant's payoff, calculated using their assigned private chip values. & \citet{Nash1950}; \citet{MurphyEtAl2011} \\
C02 & \textbf{Integrative trade search}. Tendency to construct or support packages that plausibly create gains for both sides using participant-visible evidence. & \citet{Nash1950}; \citet{CharnessRabin2002}; \citet{ChakrabortyHarbaugh2003} \\
C03 & \textbf{Preference for equal exchange quantities}. Preference for equal or nearly equal quantities of chips on both sides of an exchange, regardless of private chip values. & \citet{Guth1982}; \citet{KagelEtAl1996}; \citet{FehrSchmidt1999} \\
C04 & \textbf{Relative-outcome concern}. Concern with avoiding disadvantage or seeking an advantage relative to others, beyond absolute own surplus. & \citet{FehrSchmidt1999}; \citet{BoltonOckenfels2000}; \citet{StraubMurnighan1995} \\
C05 & \textbf{Proposal claim level}. How much the participant seeks to gain from proposed trades, given the available chips. & \citet{Nash1950}; \citet{Guth1982}; \citet{RothEtAl1991} \\
C06 & \textbf{Acceptance-calibrated offer adaptation}. Degree to which proposals become more likely to be accepted after observed responder feedback. & \citet{Rubinstein1982}; \citet{Guth1982}; \citet{StraubMurnighan1995} \\
C07 & \textbf{Package breadth}. Preference for exchanging larger multi-chip packages rather than small packages, given the available chips. & \citet{ChakrabortyHarbaugh2003} \\
C08 & \textbf{Exchange-ratio simplicity}. Preference for simple equal-count or low-complexity exchange ratios. & \citet{Guth1982}; \citet{ChakrabortyHarbaugh2003} \\
C09 & \textbf{Inventory and option-value planning}. Use of current holdings in ways that preserve or create later bargaining opportunities. & \citet{Rubinstein1982}; \citet{ChakrabortyHarbaugh2003} \\
C10 & \textbf{Own-surplus acceptance threshold}. Minimum gain the participant requires to accept an offer. & \citet{Guth1982}; \citet{StraubMurnighan1995}; \citet{RothEtAl1991} \\
C11 & \textbf{Profitable-offer uptake}. Reliability of accepting offers that clearly increase the participant's payoff under their private chip values. & \citet{Guth1982}; \citet{KagelEtAl1996} \\
C12 & \textbf{Fairness-motivated rejection}. Rejection attributable to visible unfairness or exploitation even when the participant would gain. & \citet{Guth1982}; \citet{StraubMurnighan1995}; \citet{KagelEtAl1996}; \citet{FehrSchmidt1999} \\
C13 & \textbf{Strategic rejection and delay}. Rejecting currently acceptable terms to seek better later terms, conceal information, or discipline proposers. & \citet{Rubinstein1982}; \citet{StraubMurnighan1995} \\
C14 & \textbf{Counterpart-value belief updating}. Revising offers or responses as observed proposals, acceptances, rejections, and changes in chip holdings reveal information about other participants' private chip values. & \citet{KagelEtAl1996}; \citet{ChakrabortyHarbaugh2003} \\
C15 & \textbf{Counterparty-contingent adaptation}. Adjusting later behavior according to which bargaining partner previously accepted, rejected, proposed favorable terms, or held relevant chips. & \citet{RothEtAl1991}; \citet{KagelEtAl1996} \\
C16 & \textbf{Relationship conditioning strength}. Strength of later favorable or unfavorable treatment conditional on a counterpart's prior treatment. & \citet{BoltonOckenfels2000}; \citet{CharnessRabin2002}; \citet{FalkEtAl2018} \\
C17 & \textbf{Endgame urgency}. Increase in willingness to trade, compromise, or accept modest gains as opportunities run out. & \citet{Rubinstein1982} \\
C18 & \textbf{Breakdown tolerance}. Willingness to risk ending with few or no trades rather than accept modest gains or moderate claims. & \citet{Rubinstein1982}; \citet{StraubMurnighan1995}; \citet{FalkEtAl2018} \\
C19 & \textbf{Own-payoff computation accuracy}. Accuracy in calculating payoffs using private chip values and checking whether trades are possible with the available chips. & \citet{Guth1982}; \citet{KagelEtAl1996} \\
C20 & \textbf{Feedback-based strategic adaptation}. Degree to which offers and responses change coherently following acceptance, rejection, and payoff feedback. & \citet{ErevRoth1998}; \citet{Rubinstein1982} \\
\bottomrule
\end{longtable}
\endgroup

\begingroup
\small
\setlength{\tabcolsep}{3pt}
\setlength{\LTcapwidth}{\linewidth}
\begin{longtable}{P{0.055\linewidth}P{0.59\linewidth}P{0.29\linewidth}}
\caption{Candidate behavioral features: Generalized 11--20. Literature motivates the constructs; definitions and scoring anchors are task-specific operationalizations.}\label{oa-tab:si-codebook-bounds1120}\\
\toprule
ID & Feature and definition & Literature \\
\midrule
\endfirsthead
\multicolumn{3}{l}{\textit{Continued from preceding page}}\\
\toprule
ID & Feature and definition & Literature \\
\midrule
\endhead
B01 & \textbf{Strategic iteration depth}. How strongly an observed request supports successive reasoning about the opponent's response. & \citet{AradRubinstein2012,Nagel1995,CamererHoChong2004} \\
B02 & \textbf{Opponent-belief sophistication and hedging}. How strongly an observed request supports a response to uncertainty about the opponent's choice. & \citet{GoereeLouisZhang2018,AradRubinstein2012} \\
B03 & \textbf{Payoff and incentive sensitivity}. The extent to which an observed request reflects the material payoff consequences of the stated rule. & \citet{McKelveyPalfrey1995,ManningHorton2025} \\
B04 & \textbf{Upper-endpoint salience}. The extent to which an observed request suggests attraction to the upper endpoint beyond what payoff incentives explain. & \citet{AradRubinstein2012} \\
B05 & \textbf{Payoff-computation precision}. How precisely an observed request follows a payoff-relevant best response or rule-specific target. & \citet{GoereeLouisZhang2018,McKelveyPalfrey1995} \\
B06 & \textbf{Bonus pursuit versus security}. The extent to which an observed request sacrifices secure base payoff to pursue the contingent bonus. & \citet{LiRong2018,ManningHorton2025} \\
B07 & \textbf{Instruction comprehension and adherence}. The consistency of the observed request and associated comprehension checks with the stated rules. & Feature assessing understanding of the game instructions; not a personality construct \\
\bottomrule
\end{longtable}
\endgroup

\FloatBarrier
\section{Feature selection and learning comparisons}
\label{oa-sec:feature-selection}

Eligibility for the comparison of individual features is determined entirely from the learning profiles. Public-goods features require both a numerical value and an assessment supporting its interpretation in at least 70\% of records with the relevant opportunity. In the recorded response format, the qualifying assessments are \texttt{supported}, \texttt{contradicted}, and \texttt{mixed}; these indicate evidence toward the high end, low end, or an intermediate value. Features describing responses after received punishment or reward (P18 and P22) were reserved for additional comparisons because their evidence depends on receiving those events; P30 was excluded after the semantic review described above. These restrictions leave 19 eligible features. The distinction from the descriptive numerical-value rate matters for P07, directional learning trend: 2,752 of 3,691 records contain a number (74.6\%), but only 2,065 of the 3,688 records with an opportunity contain a number with a qualifying assessment (56.0\%), below the selection threshold. Bargaining features must be identifiable in at least 50\% of learning participants, yielding 15. All seven generalized 11--20 features are eligible.

Every eligible feature is then evaluated on its own. Public goods uses the fixed panel of 48 learning games balanced across communication, punishment and reward availability. Bargaining and generalized 11--20 use the two comparisons described in Section~\ref{oa-sec:source-units}, exchanging which group provides the profiles used in simulation and which provides the human evaluation outcomes. In those two game families, the leading one-feature profile must improve on simulation without a profile in both comparisons as well as when their outcomes are combined. For this combined result, we pool outcomes across the two learning comparisons within the prespecified groups used to calculate Wasserstein distance (Appendix~\ref{oa-sec:outcomes}), then recompute the loss.

Before adding features to the leading one-feature profile, we apply a redundancy constraint to the two larger candidate feature sets. Using only profiles from the learning data, we calculate Spearman correlations using the learning records on which both features in each pair are identifiable and connect two features when $|\rho|\geq0.70$. The threshold is used to find groups of near-duplicate measurements. It does not remove features with weak correlations: every feature without a qualifying link remains eligible on its own. One representative is chosen from each connected group. Public goods chooses the representative with the best one-feature learning loss within a group. Bargaining chooses the feature with the highest identification rate. Ties are resolved first by the profile generator's mean certainty and then by feature identifier. Both rules use only learning data; the bargaining rule does not use simulation outcomes. The generalized 11--20 candidate set contains only seven features, so every unselected feature remains eligible without the redundancy constraint.

\begin{figure}[htbp]
\centering
\includegraphics[width=\textwidth]{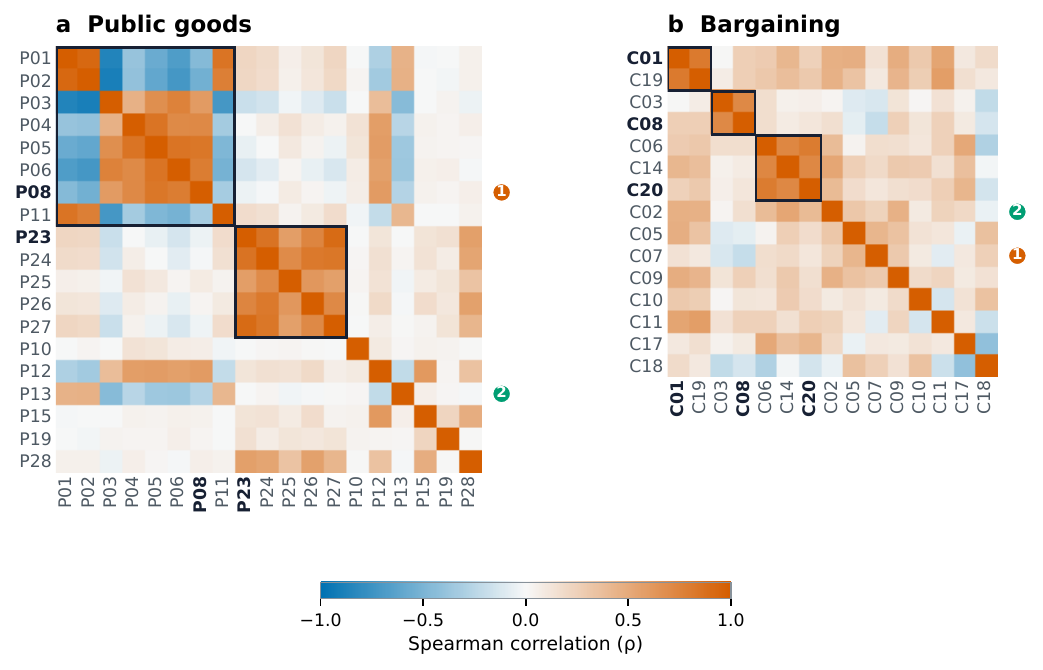}
\caption{\textbf{Correlations among learning-profile feature values define the redundancy groups.} Each cell shows a Spearman correlation calculated from records on which both features are identifiable; orange indicates positive association and blue negative association. Dark outlines enclose groups connected by links at $|\rho|\geq0.70$. Two features can belong to the same group through intermediate links, so not every pair within an outline must exceed the threshold. Bold identifiers mark the representative retained from each group. Orange markers (1) identify the leading feature and green markers (2) the second selected feature; these features form the leading two-feature profile. Feature names and definitions appear in Tables~\ref{oa-tab:si-codebook-pgg} and \ref{oa-tab:si-codebook-chip}. Generalized 11--20 is omitted because all seven features remained eligible without this constraint. Only learning records and learning comparisons enter the correlations, grouping, and feature selection.}
\label{oa-fig:si-redundancy-correlations}
\end{figure}

Figure~\ref{oa-fig:si-redundancy-correlations} shows the learning-profile correlation structure, and the following paragraph identifies their representatives. In public goods, P08 represents the contribution-response group and P23 the communication group; P10, P12, P13, P15, P19, and P28 remain eligible outside these groups. In bargaining, C01, C08, and C20 represent the three linked groups; C02, C05, C07, C09, C10, C11, C17, and C18 remain eligible outside these groups. Thus the constraint reduces the candidate additions from 19 to eight in public goods and from 15 to 11 in bargaining.

Starting with the leading one-feature profile, forward feature selection evaluates every candidate addition remaining after the redundancy constraint while keeping the same profiles assigned to the simulated participants. For current profile $A_{k-1}$ and learning loss $L_c$ in comparison $c$, an addition is eligible only when
\[
L_c(A_{k-1}\cup\{d\})<L_c(A_{k-1})
\quad\text{for every prespecified } c,
\]
and the loss calculated from the combined learning outcomes also falls. The qualifying addition with the lowest combined loss is added to the profile. Selection stops when no remaining candidate meets this rule. In public goods, withdrawal after adverse peer behavior (P08) is selected first and rule comprehension (P13) is added. In bargaining, package breadth (C07) is selected first and integrative trade search (C02) is added. In generalized 11--20, payoff and incentive sensitivity (B03) is selected first and instruction comprehension and adherence (B07) is added. No third feature meets the stopping rule in any of the three game families.

This procedure is forward feature selection with a redundancy constraint. It evaluates every eligible feature on its own and every candidate addition remaining at each step, but does not evaluate all possible subsets. It can miss features that become useful only when added jointly, and the representative of a group of correlated features need not be the member that would work best in every larger combination. Feature selection is complete before validation data are examined. Table~\ref{oa-tab:si-learning-comparisons} reports the learning comparisons for candidate additions, including the third-feature stopping tests in all three game families. None of the six public-goods additions lowers learning error relative to P08+P13. In bargaining, two of the nine additions lower the combined loss, but neither improves both learning comparisons; the remaining additions also fail at least one requirement. The five generalized 11--20 additions likewise fail to improve both learning comparisons and the combined result. The selection paths therefore stop at two features.

Public-goods addition comparisons use 45--47 games with usable outputs for both profiles when testing second-feature additions and 48 when testing third features. Bargaining second-feature comparisons use 8--14 games in each learning comparison and 21--26 across the two comparisons. Third-feature tests use 2--7 cohorts in each comparison and 8--14 across both, with both games usable for both profiles in every included cohort. Generalized 11--20 uses the fixed learning designs.

Table~\ref{oa-tab:rank-comparisons} reports every eligible one-feature comparison in both stages. The validation rankings provide a secondary check of transfer and never feed back into feature selection.

\begingroup
\small
\setlength{\tabcolsep}{3pt}
\setlength{\LTcapwidth}{\linewidth}
\begin{longtable}{P{0.15\linewidth}C{0.10\linewidth}C{0.18\linewidth}C{0.18\linewidth}C{0.13\linewidth}C{0.13\linewidth}}
\caption{Every eligible one-feature comparison. Positive error reduction means improvement over no profile. Public goods uses game-level absolute error; bargaining and generalized 11--20 use their specified Wasserstein distances. Asterisks mark features included in the primary one- or two-feature profiles. Ranks use only the stated stage.}\label{oa-tab:rank-comparisons}\\
\toprule
Family & Feature & Learning reduction & Validation reduction & Learning rank & Validation rank \\
\midrule
\endfirsthead
\multicolumn{6}{l}{\textit{Continued from preceding page}}\\
\toprule
Family & Feature & Learning reduction & Validation reduction & Learning rank & Validation rank \\
\midrule
\endhead
Public goods & P08$^{*}$ & 0.1488 & 0.0604 & 1 & 5 \\
Public goods & P06 & 0.1432 & 0.0643 & 2 & 3 \\
Public goods & P01 & 0.1415 & 0.0642 & 3 & 4 \\
Public goods & P03 & 0.1391 & 0.0667 & 4 & 1 \\
Public goods & P12 & 0.1377 & 0.0439 & 5 & 9 \\
Public goods & P02 & 0.1345 & 0.0455 & 6 & 8 \\
Public goods & P10 & 0.1321 & 0.0482 & 7 & 7 \\
Public goods & P04 & 0.1253 & 0.0656 & 8 & 2 \\
Public goods & P13$^{*}$ & 0.1218 & 0.0419 & 9 & 10 \\
Public goods & P05 & 0.0961 & 0.0579 & 10 & 6 \\
Public goods & P11 & 0.0790 & 0.0197 & 11 & 11 \\
Public goods & P28 & 0.0326 & 0.0120 & 12 & 12 \\
Public goods & P23 & 0.0053 & -0.0156 & 13 & 16 \\
Public goods & P27 & -0.0189 & -0.0394 & 14 & 17 \\
Public goods & P24 & -0.0212 & 0.0112 & 15 & 13 \\
Public goods & P15 & -0.0257 & -0.0077 & 16 & 15 \\
Public goods & P26 & -0.0260 & -0.0066 & 17 & 14 \\
Public goods & P19 & -0.0278 & -0.0413 & 18 & 18 \\
Public goods & P25 & -0.0650 & -0.0562 & 19 & 19 \\
Bargaining & C07$^{*}$ & 0.1436 & 0.1804 & 1 & 1 \\
Bargaining & C05 & 0.1259 & 0.0688 & 2 & 6 \\
Bargaining & C09 & 0.1245 & 0.0835 & 3 & 3 \\
Bargaining & C02$^{*}$ & 0.0950 & 0.0883 & 4 & 2 \\
Bargaining & C01 & 0.0928 & 0.0834 & 5 & 4 \\
Bargaining & C17 & 0.0901 & -0.0048 & 6 & 14 \\
Bargaining & C20 & 0.0828 & 0.0180 & 7 & 10 \\
Bargaining & C14 & 0.0478 & 0.0674 & 8 & 7 \\
Bargaining & C08 & 0.0419 & 0.0596 & 9 & 8 \\
Bargaining & C06 & 0.0270 & 0.0775 & 10 & 5 \\
Bargaining & C03 & 0.0143 & 0.0465 & 11 & 9 \\
Bargaining & C19 & 0.0073 & 0.0084 & 12 & 12 \\
Bargaining & C18 & -0.0077 & 0.0177 & 13 & 11 \\
Bargaining & C10 & -0.0124 & -0.0038 & 14 & 13 \\
Bargaining & C11 & -0.0699 & -0.0293 & 15 & 15 \\
11--20 & B03$^{*}$ & 0.0931 & 0.1020 & 1 & 1 \\
11--20 & B04 & 0.0699 & 0.0758 & 2 & 4 \\
11--20 & B05 & 0.0671 & 0.0893 & 3 & 2 \\
11--20 & B06 & 0.0636 & 0.0842 & 4 & 3 \\
11--20 & B01 & 0.0353 & 0.0555 & 5 & 5 \\
11--20 & B07$^{*}$ & 0.0175 & 0.0221 & 6 & 6 \\
11--20 & B02 & -0.0081 & 0.0127 & 7 & 7 \\
\bottomrule
\end{longtable}
\endgroup

\begingroup
\small
\setlength{\tabcolsep}{3pt}
\setlength{\LTcapwidth}{\linewidth}
\begin{longtable}{P{0.15\linewidth}P{0.13\linewidth}P{0.08\linewidth}C{0.12\linewidth}C{0.12\linewidth}C{0.12\linewidth}C{0.10\linewidth}}
\caption{Changes in learning error from adding one feature. Each entry is error after minus error before, calculated from the recorded comparison; negative values favor the addition. Comparisons A and B exchange which group provides profiles and which provides human evaluation outcomes. The combined column recomputes the loss using outcomes from both comparisons; for public goods, it reports the single learning panel. Every addition was tested, and no third feature qualifies. Because each addition uses observations shared with its current profile, the baseline error can differ across rows.}\label{oa-tab:si-learning-comparisons}\\
\toprule
Family & Current profile & Added feature & Comparison A & Comparison B & Combined / panel & Selected \\
\midrule
\endfirsthead
\multicolumn{7}{l}{\textit{Continued from preceding page}}\\
\toprule
Family & Current profile & Added feature & Comparison A & Comparison B & Combined / panel & Selected \\
\midrule
\endhead
Public goods & P08 & P13 & --- & --- & -0.0196 & Yes \\
Public goods & P08 & P19 & --- & --- & -0.0061 & No \\
Public goods & P08 & P15 & --- & --- & -0.0020 & No \\
Public goods & P08 & P28 & --- & --- & +0.0051 & No \\
Public goods & P08 & P23 & --- & --- & +0.0099 & No \\
Public goods & P08 & P10 & --- & --- & +0.0126 & No \\
Public goods & P08 & P12 & --- & --- & +0.0152 & No \\
Public goods & P08+P13 & P19 & --- & --- & +0.0081 & No \\
Public goods & P08+P13 & P28 & --- & --- & +0.0082 & No \\
Public goods & P08+P13 & P23 & --- & --- & +0.0124 & No \\
Public goods & P08+P13 & P15 & --- & --- & +0.0186 & No \\
Public goods & P08+P13 & P12 & --- & --- & +0.0257 & No \\
Public goods & P08+P13 & P10 & --- & --- & +0.0301 & No \\
Bargaining & C07 & C01 & +0.0748 & +0.0037 & +0.0662 & No \\
Bargaining & C07 & C02 & -0.0151 & -0.0270 & -0.0413 & Yes \\
Bargaining & C07 & C05 & +0.0185 & +0.1162 & +0.0991 & No \\
Bargaining & C07 & C08 & -0.0196 & +0.0622 & -0.0253 & No \\
Bargaining & C07 & C09 & -0.0543 & +0.1840 & +0.0375 & No \\
Bargaining & C07 & C10 & +0.1679 & +0.1165 & +0.1434 & No \\
Bargaining & C07 & C11 & +0.0339 & -0.0048 & +0.0102 & No \\
Bargaining & C07 & C17 & +0.0553 & +0.0178 & -0.0410 & No \\
Bargaining & C07 & C18 & +0.3434 & +0.0146 & +0.1103 & No \\
Bargaining & C07 & C20 & +0.0075 & +0.0039 & -0.0031 & No \\
Bargaining & C07+C02 & C01 & -0.0053 & +0.0715 & +0.0060 & No \\
Bargaining & C07+C02 & C05 & +0.0211 & +0.0124 & -0.0169 & No \\
Bargaining & C07+C02 & C08 & -0.0185 & +0.0591 & +0.0009 & No \\
Bargaining & C07+C02 & C09 & -0.0312 & +0.1780 & +0.0307 & No \\
Bargaining & C07+C02 & C10 & -0.0521 & +0.1302 & +0.0336 & No \\
Bargaining & C07+C02 & C11 & +0.0628 & +0.1234 & -0.0303 & No \\
Bargaining & C07+C02 & C17 & +0.1179 & -0.0229 & +0.0516 & No \\
Bargaining & C07+C02 & C18 & +0.0855 & +0.1086 & +0.0722 & No \\
Bargaining & C07+C02 & C20 & +0.0753 & +0.0980 & +0.0676 & No \\
11--20 & B03 & B01 & +0.0122 & -0.0021 & +0.0010 & No \\
11--20 & B03 & B02 & +0.0120 & -0.0037 & +0.0059 & No \\
11--20 & B03 & B04 & +0.0073 & -0.0049 & +0.0037 & No \\
11--20 & B03 & B05 & -0.0088 & -0.0154 & -0.0120 & No \\
11--20 & B03 & B06 & +0.0012 & -0.0048 & +0.0051 & No \\
11--20 & B03 & B07 & -0.0100 & -0.0123 & -0.0139 & Yes \\
11--20 & B03+B07 & B01 & +0.0048 & +0.0031 & +0.0060 & No \\
11--20 & B03+B07 & B02 & +0.0059 & +0.0043 & +0.0049 & No \\
11--20 & B03+B07 & B04 & -0.0003 & -0.0017 & +0.0018 & No \\
11--20 & B03+B07 & B05 & +0.0019 & -0.0155 & -0.0016 & No \\
11--20 & B03+B07 & B06 & +0.0094 & -0.0029 & +0.0097 & No \\
\bottomrule
\end{longtable}
\endgroup

\FloatBarrier
\section{Evaluation, paired comparisons and missing outputs}
\label{oa-sec:outcomes}

For comparison group $g$, let $\widehat F_{Xg}$ and $\widehat F_{Hg}$ denote simulated and human outcome distributions. The groups are defined before simulation from the rules of each human experiment: the 40 exact experimental designs in public goods, the original and alternate private-value games in bargaining, and the 11 bonus-rule families in generalized 11--20. Public goods uses normalized group efficiency, bargaining uses final surplus ratio, and generalized 11--20 uses normalized requested amount. We calculate one-dimensional Wasserstein distance within each group and average the distances equally,
\[
D(X,H)=\frac{1}{|\mathcal G|}\sum_{g\in\mathcal G}W_1(\widehat F_{Xg},\widehat F_{Hg}).
\]
Outcomes are pooled within these prespecified groups, whose definitions differ by game family. In generalized 11--20, a bonus-rule family includes multiple payoff designs; Section~\ref{oa-subsec:1120-diagnostics} examines the consequences of this aggregation. Equal weighting prevents groups with more observations from dominating the study-level measure. The normalized descriptive scale is
\[
S(X)=100\frac{D_{\mathrm{reference}}-D(X,H)}{D_{\mathrm{reference}}-D_{\mathrm{human}}}.
\]
Zero equals the expected match from a prespecified uninformed reference and 100 the estimated match between two resampled human datasets. Bargaining uses random valid actions, and generalized 11--20 uses uniformly random permitted requests. Public goods uses expected normalized efficiency of 0.5 under uniform contribution. Scores can extend below 0 or above 100.

Uncertainty is estimated by resampling games, cohorts or payoff designs, as appropriate for each game family. It is conditional on the prespecified experiments, assigned profiles and the single output generated for each requested simulation. The bootstrap intervals for each condition's alignment score in main-text Figure~\ref{fig:heldout} use 5,000 percentile-bootstrap draws. Public-goods draws resample games within the eight groups defined by the availability of communication, punishment and reward, bargaining draws resample complete cohorts and preserve their two games, and generalized 11--20 draws resample payoff designs within bonus-rule families and preserve every human choice attached to a sampled design. These eight public-goods groups govern resampling only; the full-sample loss averages the 40 exact experimental designs defined above. In each draw, distances are averaged over the exact designs represented by the sampled games. The uninformed reference and human sampling benchmark are held at their full-sample estimates when each resampled loss is transformed to the normalized scale.

The direct comparisons use the difference in loss between conditions evaluated on the same games, cohorts, or payoff designs. All three families use the Wasserstein distances underlying their normalized alignment scores. Public goods averages the distances across exact experimental designs; game-level absolute error remains the objective for feature selection and the comparison of feature rankings across stages. Each bootstrap draw resamples the same units for both conditions and recalculates their difference. We assess differences between conditions using the resulting 95\% bootstrap intervals for the paired differences in loss. A difference whose interval includes zero is inconclusive, rather than evidence of equal performance. A game or decision with an unusable output in either condition is excluded from the paired comparison. Public-goods primary, cross-family, and model-comparison paired intervals use 20,000 game-bootstrap draws within the eight groups defined above; generic-profile controls use 10,000. The same game indices select both simulated outcomes and their human reference observations in each draw. The primary bargaining comparisons use 10,000 cohort-bootstrap draws. Generalized 11--20 uses 5,000 payoff-design-bootstrap draws for primary comparisons and 10,000 for external-persona comparisons. Unusable outputs are documented in the reproducibility files and are not replaced by simulated actions.

The normalized scale provides a common reference for the three game families, while paired comparisons retain each family's outcome measure and independent unit. Table~\ref{oa-tab:alignment} reports the games or decisions with usable outputs and the independent units behind each condition in main-text Figure~\ref{fig:heldout}. Every public-goods condition displayed in main-text Figures~\ref{fig:heldout}--\ref{fig:cross-family} uses all 250 games across all 40 experimental designs, which also form the human reference. Bargaining uses all 48 games from 24 cohorts for the leading one-feature, leading two-feature, complete, and no-profile conditions; the general-purpose persona conditions contain between 44 and 47 usable games from those same 24 cohorts. All 884 generalized 11--20 payoff designs enter every displayed condition.

Table~\ref{oa-tab:si-paired-comparisons} brings together the paired comparisons with no profile, the leading two-feature and complete profiles, and the three general-purpose persona libraries. It reports the observations shared by each pair, so differences in available bargaining outputs remain explicit.

We check that each response corresponds to the intended simulation, comes from the specified model, and contains the required output in the expected format. We also check the service's completion status. Generalized 11--20 responses must contain a permitted integer; public-goods and bargaining responses must contain the actions and outcomes needed for scoring. Unusable outputs remain missing. Condition summaries use all usable observations, while paired comparisons use observations with usable outputs in both conditions. A bargaining cohort enters a paired comparison only when both games are usable in both conditions.
\begingroup
\small
\setlength{\tabcolsep}{3pt}
\setlength{\LTcapwidth}{\linewidth}
\begin{longtable}{P{0.16\textwidth}P{0.29\textwidth}P{0.06\textwidth}P{0.12\textwidth}P{0.24\textwidth}}
\caption{Paired comparisons of the leading one-feature profile with alternative profile conditions. Differences are leading one-feature minus comparison; negative values favor the leading one-feature profile. All three families use the Wasserstein distances underlying their normalized alignment scores. Public goods averages across exact experimental designs. Independent units are games, cohorts, and payoff designs, respectively. Public goods uses 20,000 bootstrap draws; bargaining uses 10,000; generalized 11--20 uses 5,000 for the primary comparisons and 10,000 for external personas. Each comparison uses the same observations in both conditions.}\label{oa-tab:si-paired-comparisons}\\
\toprule
Game family & Comparison & Units & Difference & 95\% bootstrap interval \\
\midrule
\endfirsthead
\multicolumn{5}{l}{\textit{Continued from preceding page}}\\
\toprule
Game family & Comparison & Units & Difference & 95\% bootstrap interval \\
\midrule
\endhead
Public goods & No profile & 250 & $-0.1115$ & $[-0.1334,-0.0709]$ \\
Public goods & Leading two-feature profile & 250 & $-0.0110$ & $[-0.0195,0.0107]$ \\
Public goods & Complete profile & 250 & $-0.0050$ & $[-0.0200,0.0154]$ \\
Public goods & Nemotron persona & 250 & $-0.1124$ & $[-0.1330,-0.0736]$ \\
Public goods & Twin-2K-500 persona & 250 & $-0.0787$ & $[-0.1004,-0.0474]$ \\
Public goods & ANES persona & 250 & $-0.1245$ & $[-0.1439,-0.0817]$ \\
Bargaining & No profile & 24 & $-0.1804$ & $[-0.2514,-0.0925]$ \\
Bargaining & Leading two-feature profile & 24 & $-0.0392$ & $[-0.1015,0.0223]$ \\
Bargaining & Complete profile & 24 & $-0.1192$ & $[-0.1754,-0.0464]$ \\
Bargaining & Nemotron persona & 22 & $-0.1245$ & $[-0.1771,-0.0561]$ \\
Bargaining & Twin-2K-500 persona & 23 & $-0.1713$ & $[-0.2365,-0.0897]$ \\
Bargaining & ANES persona & 20 & $-0.1392$ & $[-0.2110,-0.0485]$ \\
Generalized 11--20 & No profile & 884 & $-0.1020$ & $[-0.1155,-0.0670]$ \\
Generalized 11--20 & Leading two-feature profile & 884 & $-0.0060$ & $[-0.0214,0.0090]$ \\
Generalized 11--20 & Complete profile & 884 & $-0.0039$ & $[-0.0235,0.0158]$ \\
Generalized 11--20 & Nemotron persona & 884 & $-0.0606$ & $[-0.0734,-0.0273]$ \\
Generalized 11--20 & Twin-2K-500 persona & 884 & $-0.0621$ & $[-0.0773,-0.0326]$ \\
Generalized 11--20 & ANES persona & 884 & $-0.0981$ & $[-0.1103,-0.0653]$ \\
\bottomrule
\end{longtable}
\endgroup

\begingroup
\small
\setlength{\tabcolsep}{3pt}
\setlength{\LTcapwidth}{\linewidth}
\begin{longtable}{P{0.13\linewidth}P{0.29\linewidth}P{0.25\linewidth}C{0.11\linewidth}P{0.15\linewidth}}
\caption{Validation alignment and usable observations for the conditions in Figure~\ref{fig:heldout}. Scores use the normalization in Appendix~\ref{oa-sec:outcomes}; higher is better. Usable outputs count simulated games in public goods and bargaining and simulated decisions in generalized 11--20. Independent units are games, cohorts and payoff designs.}\label{oa-tab:alignment}\\
\toprule
Family & Condition & Alignment [95\% interval] & Usable outputs & Independent units \\
\midrule
\endfirsthead
\multicolumn{5}{l}{\textit{Continued from preceding page}}\\
\toprule
Family & Condition & Alignment [95\% interval] & Usable outputs & Independent units \\
\midrule
\endhead
Public goods & Leading one-feature profile & 65.5 [37.6, 68.1] & 250 & 250 games \\
Public goods & Leading two-feature profile & 59.0 [34.5, 65.5] & 250 & 250 games \\
Public goods & Complete profile & 62.5 [37.9, 65.2] & 250 & 250 games \\
Public goods & Nemotron persona & -1.5 [-24.7, 7.4] & 250 & 250 games \\
Public goods & Twin-2K-500 persona & 18.6 [-5.7, 23.2] & 250 & 250 games \\
Public goods & ANES persona & -8.7 [-29.4, 0.8] & 250 & 250 games \\
Public goods & No profile & -1.0 [-22.6, 7.5] & 250 & 250 games \\
Bargaining & Leading one-feature profile & 77.5 [62.8, 89.2] & 48 & 24 cohorts \\
Bargaining & Leading two-feature profile & 69.5 [53.9, 82.0] & 48 & 24 cohorts \\
Bargaining & Complete profile & 53.1 [39.9, 65.4] & 48 & 24 cohorts \\
Bargaining & Nemotron persona & 50.8 [36.9, 61.5] & 46 & 24 cohorts \\
Bargaining & Twin-2K-500 persona & 43.2 [30.5, 57.3] & 47 & 24 cohorts \\
Bargaining & ANES persona & 46.6 [33.1, 63.1] & 44 & 24 cohorts \\
Bargaining & No profile & 40.5 [29.6, 50.6] & 48 & 24 cohorts \\
11--20 & Leading one-feature profile & 73.5 [54.6, 75.9] & 884 & 884 payoff designs \\
11--20 & Leading two-feature profile & 68.0 [49.5, 73.8] & 884 & 884 payoff designs \\
11--20 & Complete profile & 70.8 [51.6, 74.0] & 884 & 884 payoff designs \\
11--20 & Nemotron persona & 37.1 [25.5, 44.5] & 884 & 884 payoff designs \\
11--20 & Twin-2K-500 persona & 36.2 [22.5, 41.3] & 884 & 884 payoff designs \\
11--20 & ANES persona & 14.6 [2.7, 22.7] & 884 & 884 payoff designs \\
11--20 & No profile & 12.2 [0.2, 20.7] & 884 & 884 payoff designs \\
\bottomrule
\end{longtable}
\endgroup

\begingroup
\small
\setlength{\tabcolsep}{3pt}
\setlength{\LTcapwidth}{\linewidth}
\begin{table}[htbp]
\centering
\caption{Full-validation normalization losses. The human sampling loss is recomputed on the usable games for cross-family bargaining conditions with missing outputs, as reported in Table~\ref{oa-tab:si-cross-domain-chip-support}.}\label{oa-tab:normalization}
\begin{tabular}{P{0.42\linewidth}C{0.24\linewidth}C{0.24\linewidth}}
\toprule
Game family & Uninformed reference & Human sampling benchmark \\
\midrule
Public goods & 0.258469438 & 0.090731881 \\
Bargaining & 0.545100531 & 0.057550213 \\
Generalized 11--20 & 0.188932147 & 0.022499968 \\
\bottomrule
\end{tabular}
\end{table}
\endgroup

\FloatBarrier
\section{General-purpose personas and cross-family comparisons}
\label{oa-sec:external-profiles}

Main-text Figure~\ref{fig:heldout} compares the selected behavioral profiles with three general-purpose persona libraries. The Nemotron condition uses rich adult records from the standard Nemotron-Personas-USA release. The Twin-2K-500 condition uses natural-language summaries of participants' survey responses. The ANES condition uses ten-statement first-person socio-demographic backstories rendered from the ANES replication data following Argyle et al. \citep{Argyle2023,ToubiaEtAl2025,NVIDIANemotron2025}. These personas describe different populations from the participants represented by the behavioral profiles.

For each general-purpose persona library, we replace the 1,946 distinct behavioral profiles used across the primary validation simulations with 1,946 distinct persona records, without using validation outcomes. Nemotron uses 1,946 distinct standard-release records, Twin-2K-500 samples without replacement from 2,058 available summaries, and ANES samples without replacement from 2,710 eligible backstories. The experimental designs, simulator, and assignment of each replacement persona to simulated participants are otherwise unchanged. Bargaining also preserves the same sampled profile for a simulated participant across the original and alternate-value games. The profile-generation and feature-selection procedure is not applied to these general-purpose persona libraries; they enter as ready-made persona text.

\subsection{Calibration of general-purpose personas using learning data}
\label{oa-subsec:external-calibration}

The comparison above evaluates the simulations without adjusting their weights to the learning outcomes. We next ask whether reweighting simulated outcomes using discrepancies observed in the learning data improves distributional alignment in validation. The persona descriptions and simulated actions remain unchanged.

For each persona library and game family, we simulate the games or decisions used in the learning comparisons. We divide the combined human and simulated learning outcomes into four ranges using their quartiles, merging boundaries when quantiles coincide. Within each range, the ratio of the human proportion to the simulated proportion determines the initial weight; adding one observation to each range before calculating these proportions smooths sparse counts. We restrict weights to $[0.25,4]$ and move them toward equal weighting when necessary to prevent a few observations from dominating. Specifically, effective sample size, $(\sum_i w_i)^2/\sum_i w_i^2$, must remain at least half the number of weighted observations. Public-goods weights are normalized within the eight groups defined by communication, punishment and reward availability. Bargaining gives both games in a cohort one joint weight, based on the product of their initial weights; generalized 11--20 estimates a separate weighting rule within each bonus-rule family.

The weighting rule estimated from learning data is then applied to the existing validation simulations. Validation outcomes do not enter its estimation, and the personas assigned to simulated participants, selected features and stopping decisions remain unchanged. Intervals in Table~\ref{oa-tab:si-learning-calibration} use 1,000 two-stage bootstrap draws that resample both the learning units used to estimate weights and the validation units used to evaluate them.

\begin{table}[htbp]
\centering
\caption{\textbf{Calibration of general-purpose persona simulations using learning data.} Lower task-specific loss is better. Change is calibrated minus equal-weight loss; negative values favor calibration. Public goods uses all 250 validation games. Bargaining uses cohorts with usable outputs for both games in each persona condition: 22 for Nemotron, 23 for Twin-2K-500, and 20 for ANES. Generalized 11--20 uses all 884 validation designs.}
\label{oa-tab:si-learning-calibration}
\small
\setlength{\tabcolsep}{4pt}
\begin{tabular}{P{0.15\textwidth}P{0.16\textwidth}C{0.14\textwidth}C{0.14\textwidth}P{0.30\textwidth}}
\toprule
Family & Persona library & Equal-weight loss & Calibrated loss & Change [95\% interval] \\
\midrule
Public goods & Nemotron & 0.2609 & 0.2571 & $-0.0039\;[-0.0121,\;0.0076]$ \\
 & Twin-2K-500 & 0.2273 & 0.2273 & $0.0000\;[-0.0095,\;0.0074]$ \\
 & ANES & 0.2731 & 0.2652 & $-0.0078\;[-0.0141,\;0.0048]$ \\
\addlinespace
Bargaining & Nemotron & 0.3078 & 0.2692 & $-0.0386\;[-0.0607,\;0.0000]$ \\
 & Twin-2K-500 & 0.3333 & 0.3211 & $-0.0121\;[-0.0401,\;0.0094]$ \\
 & ANES & 0.3162 & 0.2997 & $-0.0165\;[-0.0653,\;0.0000]$ \\
\addlinespace
Generalized 11--20 & Nemotron & 0.1272 & 0.0894 & $-0.0378\;[-0.0471,\;-0.0052]$ \\
 & Twin-2K-500 & 0.1287 & 0.0965 & $-0.0322\;[-0.0546,\;-0.0020]$ \\
 & ANES & 0.1647 & 0.1070 & $-0.0577\;[-0.0708,\;-0.0294]$ \\
\bottomrule
\end{tabular}
\end{table}

Calibration changes little in public goods: every interval includes zero. Bargaining point estimates improve by 4\%--13\%, but their intervals do not lie strictly below zero. All three persona libraries improve generalized 11--20 simulations, with loss reductions of 25\%--35\%. The calibrated loss estimates nevertheless remain above the leading one-feature profile's loss across the full validation sample in every family: 0.1486 in public goods, 0.1672 in bargaining, and 0.0666 in generalized 11--20. Only the general-purpose persona simulations are reweighted, and the usable bargaining observations differ across libraries, so this comparison also reflects those differences. Calibration improves these simulations most clearly in generalized 11--20. Even there, their point estimates do not reach the agreement achieved by the leading behavioral profile. The comparison leaves open how much of the remaining gap reflects the populations represented by the personas or the information included about them.

\subsection{Profiles constructed for another game family}
\label{oa-sec:cross-family}
For each of the six directions of cross-family transfer in Figure~\ref{fig:cross-family}, the target games, player roles, instructions, response format and GPT-5-mini simulator remain unchanged. Only the behavioral-profile portion of the prompt changes. One- and two-feature profiles contain the features selected using learning data in the family from which they were constructed; complete profiles contain every identifiable feature. No cross-family feature selection is performed.

Assignments are generated before validation outcomes are accessed, with a separate shuffle for each direction of transfer. The available pools contain 3,691 public-goods profiles, 144 bargaining profiles from the two- and three-chip designs, and 1,337 generalized 11--20 response types representing 1,670 choices. Generalized 11--20 profiles are sampled with probabilities proportional to the number of human choices they represent; other profiles have equal sampling probabilities. The shuffled sequence is cycled when necessary, but a profile is never assigned to more than one participant in the same simulated group. Assignments remain fixed across profile conditions, and bargaining retains each profile across both games.

The simulator may use a feature as uncertain background information where its meaning plausibly carries across games. It must not treat earlier actions, messages, rules or payoffs as observations from the current game or infer hidden current-game information from the profile. Scores use the target family's normalization in Appendix~\ref{oa-sec:outcomes}. Figure arrows indicate paired 95\% intervals excluding zero in the favorable or unfavorable direction. Public-goods comparisons use all 250 games and generalized 11--20 comparisons all 884 designs. The bargaining games and cohorts included are reported in Table~\ref{oa-tab:si-cross-domain-chip-support}.
\begin{table}[htbp]
\centering
\caption{\textbf{Bargaining observations for cross-family comparisons in main-text Figure~\ref{fig:cross-family}.} The human-reference loss is recomputed on each condition's usable games for the descriptive alignment score. Paired cohorts have usable outputs for both games under both the cross-family profile and no profile; they are used to estimate the interval for that comparison.}
\label{oa-tab:si-cross-domain-chip-support}
\small
\setlength{\tabcolsep}{5pt}
\begin{tabular}{P{0.21\linewidth}P{0.13\linewidth}C{0.13\linewidth}C{0.22\linewidth}C{0.18\linewidth}}
\toprule
Family used to construct profiles & Profile size & Usable games & Paired cohorts vs. no profile & $D_{\mathrm{human}}$ \\
\midrule
Public goods & One feature & 47 & 23 & 0.057963316 \\
Public goods & Two features & 46 & 22 & 0.058861741 \\
Public goods & Complete & 45 & 21 & 0.059357288 \\
Generalized 11--20 & One feature & 48 & 24 & 0.057550213 \\
Generalized 11--20 & Two features & 46 & 22 & 0.058666046 \\
Generalized 11--20 & Complete & 48 & 24 & 0.057550213 \\
\bottomrule
\end{tabular}
\end{table}

\FloatBarrier
\section{Model comparisons and sensitivity analyses}
\label{oa-sec:model-roles}
We evaluate three generator/simulator combinations in each game family: GPT-5-mini in both roles, GPT-5.1 generation with GPT-5-mini simulation, and GPT-5-mini generation with GPT-5.1 simulation. Each combination repeats the comparison of individual features and forward feature selection using only learning data before validation. GPT-5.1 in both roles was not tested. Table~\ref{oa-tab:si-model-role-matrix} reports all nine comparisons.

Here, a compact profile contains the features listed in its row of Table~\ref{oa-tab:si-model-role-matrix}. These profiles have lower estimated loss than no profile in all nine comparisons, with eight intervals excluding zero. No complete profile clearly outperforms its corresponding compact profile. The identities and number of selected features vary: alternative public-goods and bargaining model combinations retain one or two features, whereas the alternative generalized 11--20 model combinations retain three or four. Model changes therefore warrant repeating feature selection during profile learning.

All public-goods comparisons use the same 250 games, 40 designs and normalization references as the main analysis. With GPT-5-mini generation and GPT-5.1 simulation in bargaining, compact, no-profile and complete scores use 34, 37 and 34 usable games against the common 48-game human reference; each paired comparison uses nine cohorts with both games usable in both conditions.

\begingroup
\small
\setlength{\tabcolsep}{3pt}
\setlength{\LTcapwidth}{\linewidth}
\begin{longtable}{P{0.12\linewidth}P{0.10\linewidth}P{0.13\linewidth}P{0.16\linewidth}P{0.20\linewidth}P{0.20\linewidth}}
\caption{Profile-generator and simulator comparisons. The first model in each pair generates profiles; the second simulates behavior. Mini denotes GPT-5-mini and 5.1 denotes GPT-5.1. The three scores are compact / no profile / complete. Differences are compact minus comparison in Wasserstein distance, with 95\% bootstrap intervals; negative values favor the compact profile. All feature sets were selected using learning data.}\label{oa-tab:si-model-role-matrix}\\
\toprule
Family & Generator / simulator & Features & Alignment scores & Difference vs. no profile & Difference vs. complete \\
\midrule
\endfirsthead
\multicolumn{6}{l}{\textit{Continued from preceding page}}\\
\toprule
Family & Generator / simulator & Features & Alignment scores & Difference vs. no profile & Difference vs. complete \\
\midrule
\endhead
Public goods & Mini/Mini & P08 & 65.5 / -1.0 / 62.5 & $-0.1115$ [$-0.1334,-0.0709$] & $-0.0050$ [$-0.0200,0.0154$] \\
Public goods & 5.1/Mini & P06+P13 & 59.7 / -10.4 / 65.3 & $-0.1175$ [$-0.1308,-0.0742$] & $0.0094$ [$-0.0020,0.0166$] \\
Public goods & Mini/5.1 & P01 & 58.6 / -9.3 / 56.2 & $-0.1140$ [$-0.1205,-0.0687$] & $-0.0041$ [$-0.0120,0.0097$] \\
Bargaining & 5.1/Mini & C07 & 77.5 / 40.5 / 53.1 & $-0.1804$ [$-0.2514,-0.0925$] & $-0.1192$ [$-0.1754,-0.0464$] \\
Bargaining & Mini/Mini & C02+C15 & 53.4 / 40.3 / 52.4 & $-0.0735$ [$-0.1357,-0.0097$] & $-0.0048$ [$-0.0857,0.0841$] \\
Bargaining & Mini/5.1 & C08 & 95.2 / 91.6 / 103.8 & $-0.0481$ [$-0.1444,0.0611$] & $0.0092$ [$-0.0608,0.0797$] \\
11--20 & Mini/Mini & B03 & 73.5 / 12.2 / 70.8 & $-0.1020$ [$-0.1155,-0.0670$] & $-0.0039$ [$-0.0235,0.0158$] \\
11--20 & 5.1/Mini & B06+B04+\newline B07 & 69.4 / 12.2 / 64.1 & $-0.0952$ [$-0.1040,-0.0638$] & $-0.0088$ [$-0.0223,0.0097$] \\
11--20 & Mini/5.1 & B04+B05+\newline B01+B02 & 76.3 / 57.2 / 66.6 & $-0.0276$ [$-0.0479,-0.0070$] & $-0.0147$ [$-0.0321,0.0022$] \\
\bottomrule
\end{longtable}
\endgroup
\subsection{Generic demographic and length-matched profiles}
\label{oa-subsec:generic-controls}

A separate comparison asks whether generic descriptions improve simulations without behavioral evidence from the source experiments. Both controls use the same sampled Nemotron-Personas-USA records and assignments. The demographic profile includes age, sex, marital status, education, occupation, and state. The length-matched profile instead includes skills, interests, and career goals. These fields are converted to text using fixed rules, without LLM rewriting, and shortened to match the length of the leading one-feature profile's section. Length is measured in tokens, the text units counted by the model's tokenizer. The permitted difference was two tokens; the recorded sections matched exactly under the \texttt{o200k\_base} tokenizer.

We compare each control with no profile and with the leading one-feature profile under the same target environments and simulator settings. The generic descriptions concern different people from those represented by the behavioral profiles. The comparison therefore tests whether these descriptions improve simulation; it leaves open how much of the difference reflects the people represented or the information included about them.

Table~\ref{oa-tab:si-generic-controls} shows that generic descriptions can improve some simulated outcomes: a demographic profile improves bargaining relative to no profile, and a length-matched profile improves agreement with the human request distribution pooled across designs within each generalized 11--20 bonus-rule family. The leading one-feature profile nevertheless has lower task-specific loss than both generic controls in every family, with every corresponding interval excluding zero.

\begin{table}[htbp]
\centering
\small
\caption{Generic-profile controls. Differences use the Wasserstein distance underlying each family's alignment score and observations usable in both conditions; negative values favor the generic profile in the comparison with no profile, whereas positive values in the comparison with the leading profile favor the leading evidence-grounded profile. Intervals resample games, cohorts, and payoff designs, respectively. Public-goods comparisons with no profile use 247 demographic and 249 length-matched games; comparisons with the leading one-feature profile use 248 and 250. Bargaining uses 24 and 23 complete cohorts, and generalized 11--20 uses 884 payoff designs for each control. These controls describe different people from the behavioral profiles, so the comparison does not separate the effects of the people represented from the information included about them.}
\label{oa-tab:si-generic-controls}
\begin{tabular}{P{0.17\textwidth}P{0.17\textwidth}P{0.25\textwidth}P{0.25\textwidth}}
\toprule
Family & Generic profile & Generic $-$ no profile (95\% bootstrap interval) & Generic $-$ leading profile (95\% bootstrap interval) \\
\midrule
Public goods & Demographic & $0.0041$ [$-0.0087,0.0294$] & $0.1223$ [$0.0820,0.1514$] \\
Public goods & Length-matched & $-0.0198$ [$-0.0347,0.0070$] & $0.0918$ [$0.0584,0.1183$] \\
Bargaining & Demographic & $-0.0540$ [$-0.1063,-0.0026$] & $0.1264$ [$0.0362,0.1966$] \\
Bargaining & Length-matched & $-0.0171$ [$-0.0745,0.0410$] & $0.1722$ [$0.0944,0.2384$] \\
Generalized 11--20 & Demographic & $-0.0107$ [$-0.0238,0.0064$] & $0.0913$ [$0.0596,0.1046$] \\
Generalized 11--20 & Length-matched & $-0.0212$ [$-0.0338,-0.0027$] & $0.0807$ [$0.0499,0.0950$] \\
\bottomrule
\end{tabular}
\end{table}

\subsection{Bargaining with each design group in both learning and validation}
\label{oa-subsec:bargaining-interpolation}

The main analysis validates profiles under a private-value structure absent from learning. To examine whether profile selection changes when learning and validation instead contain similar mixtures of designs, a second prespecified split assigns 12 of 24 cohorts in each of the two-, three- and four-chip design groups to learning and 12 to validation, keeping both games and all participants together. Within this split, each learning comparison and the validation comparison use a prespecified panel of 18 games.

The equal split selects C05, proposal claim level. In its validation panel, C05 scores 67.8, compared with 52.5 for no profile. The complete profile scores 81.0. Profiles restricted to features meeting the identification threshold score 73.0; additionally applying the redundancy constraint yields 76.0. C05's paired improvement is imprecise (95\% interval, $-0.1040$ to 0.0743 in task-specific loss), whereas the complete profile improves clearly on no profile (interval, $-0.2394$ to $-0.0705$). No second feature improves C05 in both learning comparisons, so this split has no leading two-feature profile to validate. The complete profile's clearer advantage in this split illustrates how the usefulness of a compact profile can depend on the difference between the learning and validation designs.

\subsection{Generalized 11--20 aggregation and action-level comparisons}
\label{oa-subsec:1120-diagnostics}

To examine whether the improvement in pooled request distributions also appears within individual payoff designs, we evaluate the same simulated requests in two additional ways. First, we measure the Wasserstein distance between the simulated request and the human request distribution within each design, averaging equally across designs within each bonus-rule family and then across families. Second, we calculate a log loss after assigning 99\% of the probability to the simulated request and spreading the remaining 1\% uniformly across permitted requests. This second comparison averages across human choices, so designs with more responses receive more weight. The probability distribution is imposed for scoring; it is not a probability forecast returned by the model. Neither comparison estimates a full simulated choice distribution within each design. Lower values indicate closer agreement in both columns of Table~\ref{oa-tab:si-bounds1120-design-diagnostics}.

\begin{table}[htbp]
\centering
\small
\caption{Generalized 11--20 action-level comparisons by payoff design. The leading feature is payoff and incentive sensitivity; the leading two-feature profile adds instruction comprehension and adherence. Design-level Wasserstein distance is averaged equally across designs within bonus-rule family and then equally across families. Log loss averages across the 2,502 observed human choices, assigning 99\% probability to the simulated request and spreading 1\% uniformly across permitted requests. No uncertainty intervals are attached to these secondary comparisons. Lower values indicate better agreement.}
\label{oa-tab:si-bounds1120-design-diagnostics}
\begin{tabular}{P{0.34\textwidth}C{0.28\textwidth}C{0.28\textwidth}}
\toprule
Profile condition & Design-level Wasserstein distance & Log loss based on the simulated request \\
\midrule
No profile & 0.2711 & 4.9486 \\
Leading one-feature profile & 0.3194 & 5.4579 \\
Leading two-feature profile & 0.3239 & 5.3828 \\
Complete profile & 0.3203 & 5.4259 \\
\bottomrule
\end{tabular}
\end{table}

Simulation without a profile has lower error on both comparisons. The improvement from behavioral profiles therefore concerns request distributions pooled across designs; it does not extend to these comparisons within individual designs.

\end{document}